\documentclass{article}

\usepackage[final]{corl2020} 
\usepackage{graphicx}
\usepackage{float}
\usepackage{amsmath}
\usepackage{amssymb}
\usepackage{url}
\usepackage{bbm}
\usepackage{wrapfig}
\usepackage[font=footnotesize]{caption}
\usepackage{makecell}
\usepackage{tabularx,multirow}
\usepackage{adjustbox}

\usepackage{scalerel}

\usepackage{algorithm}
\usepackage{algpseudocode}
\algnewcommand{\Comment}[1]{\hfill// \eqparbox{COMMENT\thealgorithm}{#1}}
\usepackage{enumitem}

\newcommand{\PAC}{\textup{PAC}}
\newcommand{\KL}{\textup{KL}}

\newcommand{\normal}{\mathcal{N}}

\newtheorem{theorem}{\bf Theorem}

\title{Generalization Guarantees for Imitation Learning}

\author{
  Allen Z. Ren, \ Sushant Veer, \ and Anirudha Majumdar\\
  Department of Mechanical and Aerospace Engineering\\
  Princeton University \\
  \texttt{\{allen.ren, sveer, ani.majumdar\}@princeton.edu}
}

\begin{document}
\maketitle



\begin{abstract}
Control policies from imitation learning can often fail to generalize to novel environments due to imperfect demonstrations or the inability of imitation learning algorithms to accurately infer the expert’s policies. In this paper, we present rigorous generalization guarantees for imitation learning by leveraging the \emph{Probably Approximately Correct (PAC)-Bayes} framework to provide upper bounds on the expected cost of policies in novel environments. We propose a two-stage training method where a latent policy distribution is first embedded with multi-modal expert behavior using a conditional variational autoencoder, and then ``fine-tuned'' in new training environments to explicitly optimize the generalization bound. We demonstrate strong generalization bounds and their tightness relative to empirical performance in simulation for (i) grasping diverse mugs, (ii) planar pushing with visual feedback, and (iii) vision-based indoor navigation, as well as through hardware experiments for the two manipulation tasks. \footnote{Code is available at: \url{https://github.com/irom-lab/PAC-Imitation}} \footnote{A video showing the experiment results is available at: \url{https://youtu.be/dfXyHvOTolc}} \footnote{A presentation video is available at: \url{https://youtu.be/nabtvOWoIlo}}
\end{abstract}


\section{Introduction}
\label{sec:intro}
\vspace{-5pt}

Imagine a personal robot that is trained to navigate around homes and manipulate objects via \emph{imitation learning} \citep{osa2018algorithmic}. How can we guarantee that the resulting control policy will behave safely and perform well when deployed in a novel environment (e.g., in a previously unseen home or with new furniture)? Unfortunately, state-of-the-art imitation learning techniques do not provide any guarantees on \emph{generalization} to novel environments and can fail dramatically when operating conditions are different from ones seen during training \citep{zhu2018reinforcement}. This may be due to the expert’s demonstrations not being safe or generalizable, or due to the imitation learning algorithm not accurately inferring the expert’s policy. The goal of this work is to address this challenge and propose a framework that allows us to provide \emph{rigorous guarantees on generalization} for imitation learning.

The key idea behind our approach is to leverage powerful techniques from \emph{generalization theory} \cite{vapnik1999overview} in theoretical machine learning to ``fine-tune" a policy learned from demonstrations while also making guarantees on generalization for the resulting policy. More specifically, we employ \emph{Probably Approximately Correct (PAC)-Bayes} theory \cite{mcallester1999some, Seeger02, langford2003pac}; PAC-Bayes theory has recently emerged as a promising candidate for providing strong generalization bounds for neural networks in supervised learning problems \cite{Dziugaite17, Neyshabur17, Arora18} (in contrast to other generalization frameworks, which often provide vacuous bounds \cite{Dziugaite17}). However, the use of PAC-Bayes theory beyond supervised learning settings has been limited. In this work, we demonstrate that PAC-Bayes theory affords previously untapped potential for providing guarantees on imitation-learned policies deployed in novel environments.

\begin{figure}[t]
\centering
\includegraphics[width=0.98\textwidth]{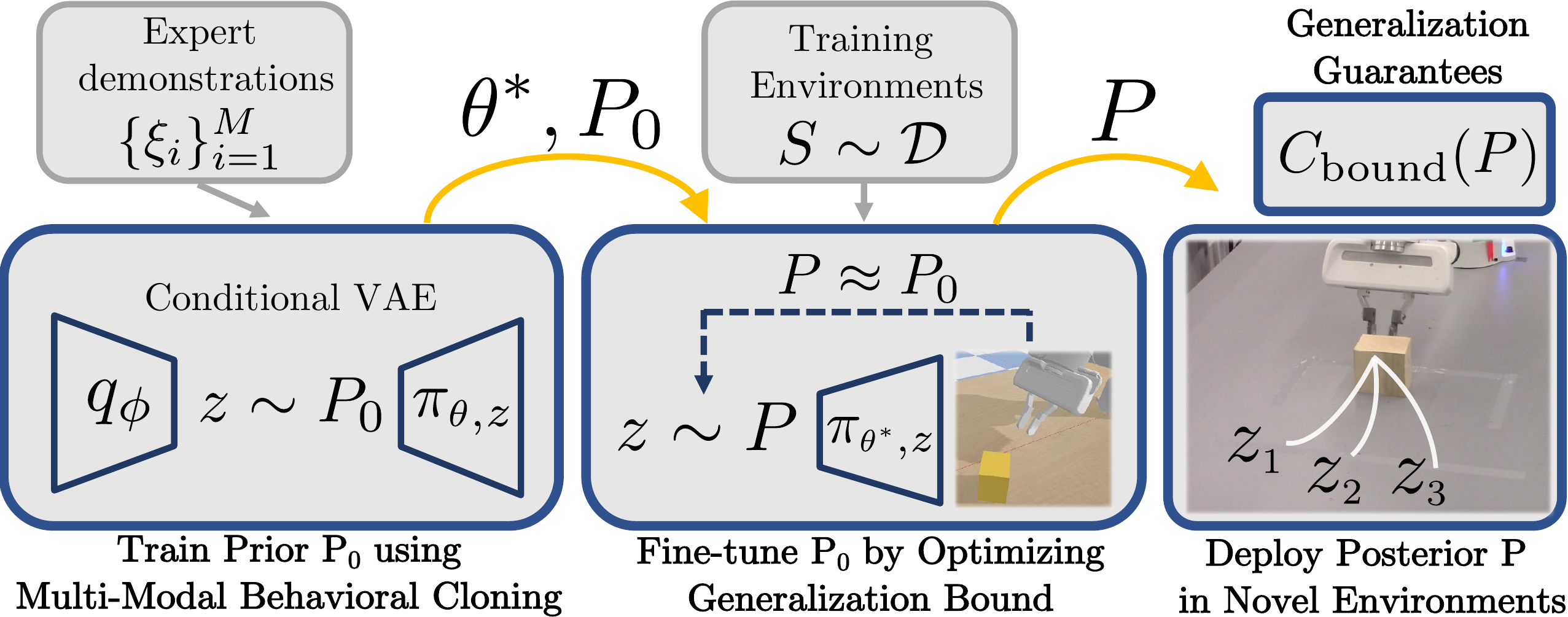}
\caption{\small System overview. (Left) policies from expert demonstrations are embedded into a ``prior'' distribution $P_0$ (latent distribution in a cVAE). The cVAE decoder $\pi_{\scaleto{\theta}{3pt},z}$ learns to generate actions close to an expert's; (middle) the prior $P_0$ is then ``fine-tuned'' by optimizing a generalization bound from PAC-Bayes theory. The decoder $\pi_{\scaleto{\theta}{3pt},z}$ (with neural network parameters fixed as $\theta^*$) is re-used to infer actions in training environments. The posterior distribution $P$ is close to $P_0$, (right) and comes with a generalization bound $C_\textup{bound}(P)$ when $P$ is deployed in novel environments. Note that we work with a distribution over policies: different $z$ sampled from $P$ produce different pushing trajectories given the same observation.}
\label{fig:anchor}
\vspace{-10pt}
\end{figure}

\emph{Statement of Contributions:} To our knowledge, the results in this paper constitute the first attempt to provide generalization guarantees on policies learned via imitation learning for robotic systems with rich sensory inputs (e.g., RGB-D images), complicated (e.g., nonlinear and hybrid) dynamics, and neural network-based policy architectures. We present a synergistic two-tier training pipeline that performs imitation learning in the first phase and then ``fine-tunes" the resulting policy in a second phase (Fig.\ \ref{fig:anchor}). In particular, the first phase performs multi-modal behavioral cloning \citep{morton2017simultaneous} using a conditional variational autoencoder (cVAE) \cite{sohn2015learning, kingma2014semi} and diverse expert demonstrations. The resulting policy is then used as a \emph{prior} for the second phase; this prior is specified by the distribution over the cVAE's latent variables. The second phase of training uses a fresh set of environments to optimize a \emph{posterior} distribution over the latent variables by explicitly optimizing a bound on the generalization performance derived from PAC-Bayes theory. The resulting fine-tuned policy has an associated bound on the expected cost across novel environments under the assumption that novel environments and training environments are drawn from the same (but \emph{unknown}) distribution (see Sec.~\ref{sec:problem formulation} for a precise problem formulation). We demonstrate strong generalization bounds and their tightness relative to empirical generalization performance using simulation experiments in three settings: (i) grasping mugs with varying geometric and physical properties, (ii) vision-based feedback control for planar pushing, and (iii) navigating in cluttered home environments. We also present extensive hardware experiments for the manipulation examples using a Franka Panda arm equipped with an external RGB-D sensor. Taken together, our simulation and hardware experiments demonstrate the ability of our approach to provide strong generalization guarantees for policies learned via imitation learning in challenging robotics settings.

\vspace{-10pt}
\subsection*{Related work}
\vspace{-5pt}
\textbf{Multi-modal imitation learning.} Imitation learning is commonly used in manipulation \cite{florence2019self, Zhang18, zhu2018reinforcement} and navigation tasks \citep{bansal2018chauffeurnet, codevilla2018end} to accelerate training by injecting expert knowledge. Often, imitation data can be multi-modal, e.g., an expert may choose to grasp anywhere along the rim of a mug. Recently, a large body of work uses latent variable models to capture such multi-modality in robotic settings \citep{hausman2017multi, wang2017robust, rahmatizadeh2018vision, hsiao2019learning, morton2017simultaneous}. While these papers use multi-modal data to diversify imitated behavior, we embed the multi-modality into the prior policy distribution to accelerate the ``fine-tuning'' and produce better generalization bounds and empirical performance (see Sec.~\ref{subsec:navigate} and \ref{sup:benefit}). 

\textbf{Learning from imperfect demonstrations.} Another challenge with imitation learning is that expert demonstrations can often be imperfect \cite{gao2018reinforcement}. Moreover, policies trained using only off-policy data can also cause cascading errors \citep{ross2011reduction} when the robot encounters unseen states. One approach to addressing these challenges is to ``fine-tune'' an imitation-learned policy to improve empirical performance in novel environments \citep{gao2018reinforcement, gupta2019relay}. Other techniques that mitigate these issues include collecting demonstrations with noisy dynamics \citep{laskey2017dart}, augmenting observation-action data of demonstrations using noise \cite{florence2019self}, and training using a hybrid reward for both imitation and task success \cite{zhu2018reinforcement}. Some recent work also explores generalization in longer horizon tasks \cite{gupta2019relay, mandlekar2020learning}. However, none of these techniques provide rigorous generalization guarantees for imitated policies deployed in novel environments for robotic systems with discrete/continuous state and action spaces, nonlinear/hybrid dynamics, and rich sensing (e.g., vision). This is the primary focus of this paper.

\textbf{Generalization guarantees for learning-based control.} The PAC-Bayes Control framework \citep{Majumdar18, Majumdar19, Veer20} provides a way to make safety and generalization guarantees for learning-based control. In this work, we extend this framework to make guarantees on policies learned via imitation learning. To this end, we propose a two-phase training pipeline for learning a prior distribution over policies via imitation learning and then fine-tuning this prior by optimizing a PAC-Bayes generalization bound. By leveraging multi-modal expert demonstrations, we are able to obtain significantly stronger generalization guarantees than \cite{Majumdar19, Veer20}, which either employ simple heuristics to choose the prior or train the prior from scratch.

\vspace{-5pt}
\section{Problem Formulation}
\label{sec:problem formulation}
\vspace{-5pt}
We assume that the discrete-time dynamics of the robot are given by:
\begin{equation}
    s_{t+1} = f_E(s_t, a_t),
\end{equation}
where $s_t \in \mathcal{S} \subseteq \mathbb{R}^{n_s}$  is the state at time-step $t$, $a_t \in \mathcal{A} \subseteq \mathbb{R}^{n_a}$ is the action, and $E \in \mathcal{E}$ is the environment that the robot is operating in. We use the term ``environment" here broadly to refer to external factors such as the object that a manipulator is trying to grasp, or a room that a personal robot is operating in. We assume that the robot has a sensor which provides observations $o_t \in \mathcal{O}$. For the first phase of our training pipeline (Fig. \ref{fig:anchor}), we assume that we are provided a finite set of expert demonstrations $\{\zeta_i\}_{i=1}^M$, where each $\zeta_i:=\{(o_t,a_t)\}_{t=1}^T$ is a sequence of observation-action pairs (e.g., human demonstrations of a manipulation task, where $o_t$ and $a_t$ correspond to depth images and desired relative-to-current poses respectively at step $t$ of the sequence). For the second phase of training, we assume access to a dataset $S := \{E_1, \dots, E_N\}$ of $N$ training environments drawn independently from a distribution $\mathcal{D}$ (e.g., the distribution on the shapes, dimensions, and mass of mugs to be grasped). Importantly, we \emph{do not} assume knowledge of the distribution $\mathcal{D}$ or the space $\mathcal{E}$ of environments (which may be extremely high-dimensional). We allow $\mathcal{D}$ to differ from the distribution from which environments for $\{\zeta_i\}_{i=1}^M$ are drawn.

Suppose that the robot's task is specified via a cost function and let $C(\pi; E)$ denote the cost incurred by a (deterministic) policy $\pi: \mathcal{O} \rightarrow \mathcal{A}$ when deployed in an environment $E$. Here, we assume that policy $\pi$ belongs to a space  $\Pi$ of policies. We also allow policies that map \emph{histories} of observations to actions by augmenting the observation space to keep track of observation sequences. The cost function is assumed to be bounded; without further loss of generality, we assume $C(\pi; E) \in [0,1]$. We make no further assumptions on the cost function (e.g., continuity, smoothness, etc.).

{\bf Goal.} Our goal is to utilize the expert demonstrations $\{\zeta_i\}_{i=1}^M$ along with the additional training environments in $S$ to learn a policy that \emph{provably generalizes} to novel environments $S'$ drawn from the \emph{unknown} distribution $\mathcal{D}$. In this work, we will employ a slightly more general formulation where we choose a \emph{distribution} $P$ over policies $\pi\in\Pi$ (instead of making a single deterministic choice). This will allow us to employ PAC-Bayes generalization theory. Our goal can then be formalized by the following optimization problem: 
\begin{flalign}
\centering
C^\star := \underset{P \in \mathcal{P}}{\textrm{min}} ~C_\mathcal{D}(P), \  \ \text{where}  \   \ C_\mathcal{D}(P) := \underset{E \sim \mathcal{D}}{\mathbb{E}} \ \underset{\pi \sim P}{\mathbb{E}} [C(\pi; E)], \label{eq:OPT}
\end{flalign}
and $\mathcal{P}$ refers to the space of probability distributions on the policy space $\Pi$. This optimization problem is challenging to tackle directly since the distribution $\mathcal{D}$ from which environments are drawn is not known to us. In the subsequent sections, we will demonstrate how to learn a distribution $P$ over policies with a provable bound on the expected cost $C_\mathcal{D}(P)$, i.e., a provable guarantee on generalization to novel environments drawn from $\mathcal{D}$.
\vspace{-5pt}
\section{Approach}
\vspace{-5pt}
The training pipeline consists of two stages (Fig.~\ref{fig:anchor}). First, a ``prior'' distribution over policies $P_0$ is obtained by cloning multi-modal expert demonstrations. Second, the prior is ``fine-tuned'' by explicitly optimizing the PAC-Bayes generalization bound. The resulting ``posterior'' policy distribution $P$ achieves strong empirical performance and generalization guarantees on novel environments.
\subsection{Multi-Modal Behavioral Cloning using Latent Variables}
\label{sec:behavior-cloning}
\begin{wrapfigure}[12]{r}{0.22\textwidth}
\vspace{-15pt}
\begin{center}
\includegraphics[width=0.22\textwidth]{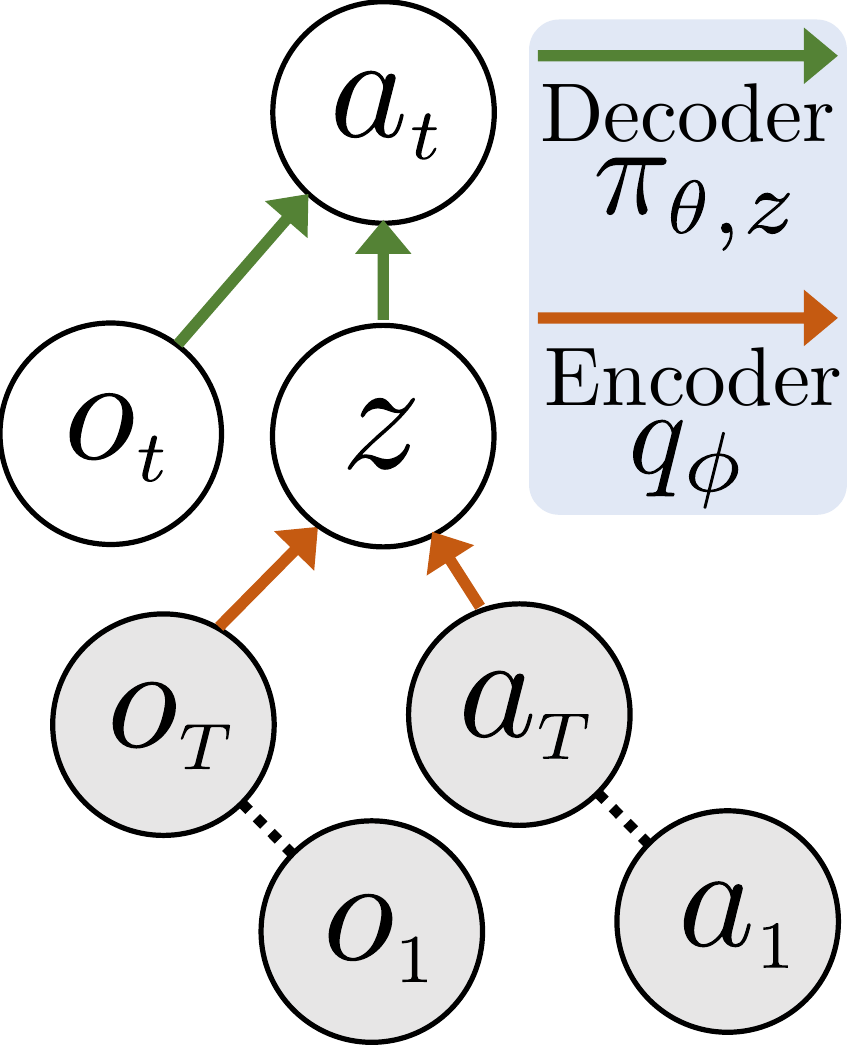}
\end{center}
\vspace{-8pt}
\caption{Graphical model of the cVAE}
\label{fig:graphical_model}
\end{wrapfigure}

\vspace{-5pt}
The goal of the first training stage is to obtain a prior policy distribution $P_0$ by cloning expert demonstrations $\zeta_i$. Behavioral cloning is a straightforward strategy to make robots mimic expert behavior. While simple discriminative models fail to capture diverse expert behavior, generative models such as variational autoencoders (VAEs) can embed such multi-modality in latent variables \citep{sohn2015learning, wang2017robust}. We further use a conditional VAE (cVAE) \citep{sohn2015learning} to condition the action output $a$ on both the latent $z$ and observation $o$ (Fig.~\ref{fig:graphical_model}). The latent $z$ encodes both $o$ and $a$ from expert demonstrations $\{\zeta_i\}_{i=1}^M$. In the example of grasping mugs, intuitively we can consider that $z$ encodes the mug center location and the relative-to-center grasp pose from depth images and demonstrated grasps. Both pieces of information are necessary for generating successful, diverse grasps along the rim using different sampled $z$ (Fig.~\ref{fig:grasp-push-samples}(a)). 

While grasping mugs can be achieved by executing an open-loop grasp from above, other tasks such as pushing boxes and navigating indoor environments require continuous, closed-loop actions. We embed either a short sequence of observation/action pairs ($T=3$ steps) or the entire trajectory (as many as 50 steps) into a single latent $z$. Thus a sampled latent state can represent local, reactive actions (e.g. turn left to avoid a chair during navigation), or global ``plans'' (e.g. push at the corner of an object throughout the horizon while manipulating it). 

Fig.~\ref{fig:graphical_model} shows the graphical model of the cVAE in our work. The encoder $q_\phi(z|o,a)$, parameterized by weights $\phi$ of a neural network, samples a latent variable $z$ conditioned by each demonstration $\zeta_i:=\{(o_t,a_t)\}_{t=1}^T$. The decoder, $\pi_{\theta,z}:o \mapsto a$, parameterized by the weights $\theta$ of another neural network and the sampled latent variable $z$, reconstructs the action from the observation at each step. Details of the neural network architecture are provided in \ref{sup:synthetic-training}. In the loss function below, the distribution $p(z)$ is chosen as a multivariate unit Gaussian. A KL regularization loss constrains the conditional distribution of $z$ to be close to $p(z)$. Let $\mathcal{L}_{\rm rec}(\pi_{\scaleto{\theta}{4pt},z}(o_t),a_t)$ be the reconstruction loss between the predicted action and the expert's. The parameter $\lambda>0$ balances the two losses:
\begin{equation}
\label{eq:cvae_loss}
\mathcal{L}(\phi,\theta;\zeta_i)= \mathrm{E}_{z \sim q_{\phi}(z | o_{1:T}, a_{1:T})}\bigg[\sum_{t=1}^T \mathcal{L}_{\rm rec}\big(\pi_{\scaleto{\theta}{4pt},z}(o_t),a_t\big)\bigg] +\lambda D_{\KL}\big((q_{\phi}(z | o_{1:T}, a_{1:T})|| p(z)\big).
\end{equation}
The primary outcomes of behavioral cloning are: (i) a latent distribution $p(z) = \normal(0,I)$ and (ii) weights $\theta^*$ of the decoder network $\pi_{\scaleto{\theta}{4pt},z}$ that together encode multi-modal policies from experts. We now restrict the weights $\theta$ of the decoder network $\pi_{\scaleto{\theta}{4pt},z}$ to $\theta^*$ giving rise to the space of policies $\Pi:=\{\pi_{\scaleto{\theta^*}{4pt},z}:\mathcal{O}\to\mathcal{A}~|~z\in\mathbb{R}^{n_z} \}$ parameterized by $z$. Hence, the latent distribution $p(z)$ can be equivalently viewed as a distribution on the space $\Pi$ of policies.  
In the next section, we will consider $p(z)$ as a \emph{prior} distribution $P_0$ on $\Pi$ and ``fine-tune" it by searching for a \emph{posterior} distribution $P$ in the space $\mathcal{P}$ of probability distributions on $\Pi$ by solving \eqref{eq:OPT}. In particular, we choose $\mathcal{P}$ as the space of Gaussian probability distributions with diagonal covariance $\normal(\mu,\Sigma)$. For the sake of notational convenience, let $\sigma \in \mathbb{R}^d$ be the element-wise square-root of the diagonal of $\Sigma$, and define $\psi :=(\mu, \sigma)$, $P = \normal_{\psi} := \normal(\mu, \text{diag}(\sigma^2))$, and $P_0 = \normal_{\psi_0} := \normal(0,I)$.

\subsection{Generalization Guarantees through PAC-Bayes Control}
\label{sec:pac-es}
\vspace{-5pt}
Although behavioral cloning provides a meaningful policy distribution $P_0$, the policies drawn from this distribution can fail when deployed in novel environments due to: unsafe or non-generalizable demonstrations by the expert, or the inability of the cVAE training to accurately infer the expert’s policies. In this section, we leverage the PAC-Bayes Control framework introduced in \citep{Majumdar18, Majumdar19} to ``fine-tune'' the \emph{prior} policy distribution $P_0$ and provide ``certificates" of generalization for the resulting \emph{posterior} policy distribution $P$. In particular, we will tune the distribution by approximately minimizing the true expected cost $C_\mathcal{D}(P)$ in equation \eqref{eq:OPT}, thus promoting generalization to environments drawn from $\mathcal{D}$ that are different from $S$. Although $C_\mathcal{D}(P)$ cannot be computed due to the lack of an explicit characterization of $\mathcal{D}$, the PAC-Bayes framework allows us to obtain an upper bound $C_\PAC$ for $C_\mathcal{D}(P)$, which can be computed despite our lack of knowledge of $\mathcal{D}$; see Theorem~\ref{thm:pac bayes control}.

To introduce the PAC-Bayes generalization bounds, we will first define the \emph{empirical cost} of $P$ as the average expected cost across training environments in $S$:
\begin{equation}
\label{eq:training_cost}
C_{S}(P):=\frac{1}{N} \sum_{E \in S} \underset{z \sim P}{\mathbb{E}}[C(\pi_{\scaleto{\theta^*}{4pt},z} ; E)].
\end{equation}
The following theorem can then be used to bound the true expected cost $C_\mathcal{D}(P)$ from Sec.~\ref{sec:problem formulation}.
\vspace{-5pt}

\begin{theorem}[PAC-Bayes Bound for Control Policies; adapted from \citep{Majumdar18,maurer2004note}] 
\label{thm:pac bayes control}
Let $P_0\in\mathcal{P}$ be a prior distribution. Then, for any $P\in\mathcal{P}$, and any $\delta \in (0,1)$, with probability at least $1 - \delta$ over sampled environments $S \sim \mathcal{D}^N$, the following inequality holds:
\vspace{-5pt}
\begin{equation*}
C_\mathcal{D}(P) \leq C_{\PAC}(P, P_0) := C_S(P) + \sqrt{R(P, P_0)}
\label{eq:maurer-pac-bound}, \ \textup{where} \ R(P,P_0) := \frac{\KL(P \| P_0) + \log (\frac{2\sqrt{N}}{\delta})}{2N}.
\end{equation*}
\vspace{-15pt}
\end{theorem}
Intuitively, minimizing the upper bound $C_\PAC$ can be viewed as minimizing the empirical cost $C_S(P)$ along with a regularizer $R$ that prevents overfitting by penalizing the deviation of the posterior from the prior. Due to the presence of a ``blackbox" physics simulator for rollouts, the gradient of $C_S(P)$ cannot be computed analytically. Thus we employ blackbox optimizers for minimizing $C_\PAC$.

\textbf{Optimizing PAC-Bayes bound using Natural Evolutionary Strategies.} To minimize $C_\PAC$, we use the class of blackbox optimizers known as Evolutionary Strategies (ES) \citep{beyer2002evolution} that estimate the gradient of a loss function through Monte-Carlo methods, without requiring an analytical gradient of the loss function. To minimize $C_\PAC$ using ES, we express the gradient of the empirical cost $C_S(P)$  \eqref{eq:training_cost} as an expectation w.r.t. the posterior distribution $P = \mathcal{N}_\psi$:
\vspace{-2pt}
\begin{equation}\label{eq:emp-ES-grad}
    \nabla_\psi C_S(\normal_\psi) = \frac{1}{N} \sum_{E \in S} \nabla_\psi \underset{z \sim \normal_\psi}{\mathbb{E}}[C(\pi_{\scaleto{\theta^*}{4pt},z} ; E)] = \frac{1}{N} \sum_{E \in S}\mathop{\mathbb{E}}_{z \sim \normal_\psi}[C(\pi_{\scaleto{\theta^*}{4pt},z};E)\nabla_\psi \ln \normal_{\psi}(z)].
\vspace{-1pt}
\end{equation}
Although the gradient of the regularizer $R$ can be computed analytically, we found that it can heavily dominate the noisy gradient estimate of the empirical cost during training. Therefore, we compute its gradient using ES as well, expressing the regularizer in terms of an expectation on the posterior:
\begin{equation}\label{eq:reg-ES-grad}
    \nabla_\psi R = \frac{1}{2N}\nabla_\psi \KL(\normal_\psi \| \normal_{\psi_0}) = \frac{1}{2N}\nabla_\psi \mathop{\mathbb{E}}_{z \sim \normal_\psi}\Big[\log\frac{\normal_\psi(z)}{\normal_{\psi_0}(z)}\Big].
\end{equation}
\begin{wrapfigure}[11]{l}{0.45\textwidth}
\vspace{-15pt}
\begin{center}
\includegraphics[width=0.35\textwidth]{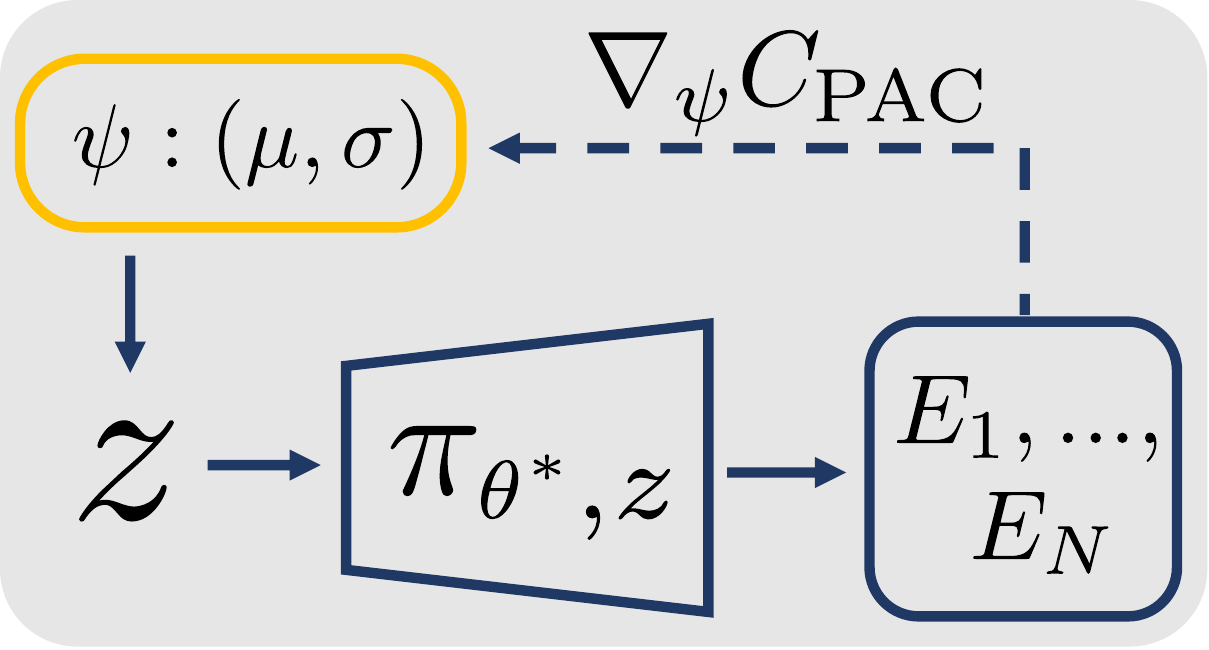}
\end{center}
\vspace{-10pt}
\caption{PAC-Bayes Control training loop. Each training environment $E_i$ is simulated with a random policy $z$ sampled from $\normal_\psi$, which is then updated using gradient estimate of the bound.}
\label{fig:pac-es}
\end{wrapfigure}
In practice we use Natural Evolutionary Strategies (NES) \citep{wierstra2014natural} that transforms the ES gradient to the natural gradient \citep{amari1998natural} to accelerate training.  During each epoch, for each of the $N$ environments we sample a certain number of $z$'s from the posterior $\mathcal{N}_\psi$ and then compute the corresponding empirical costs by performing rollouts in simulation. Sampled $z$'s and their empirical costs $C(\pi_{\scaleto{\theta^*}{4pt},z};E)$ are then used in \eqref{eq:emp-ES-grad} and \eqref{eq:reg-ES-grad} to compute the gradient estimate, which is passed to the Adam optimizer \citep{kingma2014adam} to update $\psi$. The training loop is visualized in Fig.~\ref{fig:pac-es}.

\textbf{Computing the final bound.} After ES training, we can calculate the generalization bound using the optimal $\psi^*$. First, note that the empirical cost $C_S(P) = C_S(\normal_{\psi^*})$ involves an expectation over the posterior and thus cannot be computed in closed form. Instead, it can be estimated by sampling a large number of policies $z_1, \cdots, z_L$ from $\normal_{\psi^*} : \hat{C}_S(\normal_\psi^*) := \frac{1}{NL}\sum_{E \in S} \sum_{i=1}^L C(\pi_{\scaleto{\theta^*}{4pt},z_i}; E)$, and the error due to finite sampling can be bounded using a sample convergence bound $\bar{C}_S$ \citep{langford2002not}. The final bound $C_\textup{bound}(\normal_\psi^*) \geq C_\mathcal{D}(\normal_\psi^*)$ is obtained from $\bar{C}_S$ and $R(\normal_\psi^*, P_0)$ by a slight tightening of $C_\PAC$ from Theorem~\ref{thm:pac bayes control} using the KL-inverse function \cite{Majumdar19}. Please refer to \ref{sup:es}, \ref{sup:final-bound}, \ref{sup:full-algorithm} for detailed derivations and implementations.

Overall, our approach provides generalization guarantees in novel environments for policies learned from imitation learning: as policies are randomly sampled from the posterior $\normal_\psi^*$ and applied in test environments, the expected success rate over all test environments is guaranteed to be at least $1-C_{\textup{bound}}(\normal_\psi^*)$ (with probability $1-\delta$ over the sampling of training environments; $\delta=0.01$ for all examples in Sec.~\ref{sec:experiments}).

\vspace{-5pt}
\section{Experiments}
\label{sec:experiments}
\vspace{-5pt}
We demonstrate the efficacy of our approach on three different robotic tasks: grasping a diverse set of mugs, planar box pushing with external vision feedback, and navigating in home environments with onboard vision feedback. Our experimental results demonstrate: (i) strong theoretical generalization bounds, (ii) tightness between theoretical bounds and empirical performance in test environments, and (iii) zero-shot generalization to hardware in challenging manipulation settings.

Expert demonstrations are collected in the PyBullet simulator \citep{coumans2019} using a 3Dconnexion 6-dof mouse for manipulation and a keyboard for navigation; no data from the real robot or camera is used for the training. Following behavioral cloning using collected data, we fine-tune the policies using rollout costs in the PyBullet simulator as well. Results of manipulation tasks are then transferred to the real hardware with no additional training (zero-shot). More details of synthetic training (including code) and hardware experiments (including a video) are provided in \ref{sup:synthetic-training} and \ref{sup:hardware-experiments}.
\vspace{-5pt}
\subsection{Grasping a diverse set of mugs}
\label{subsec:grasp}
\begin{wrapfigure}[12]{r}{0.40\textwidth}
\vspace{-16pt}
\begin{center}
\includegraphics[width=0.40\textwidth]{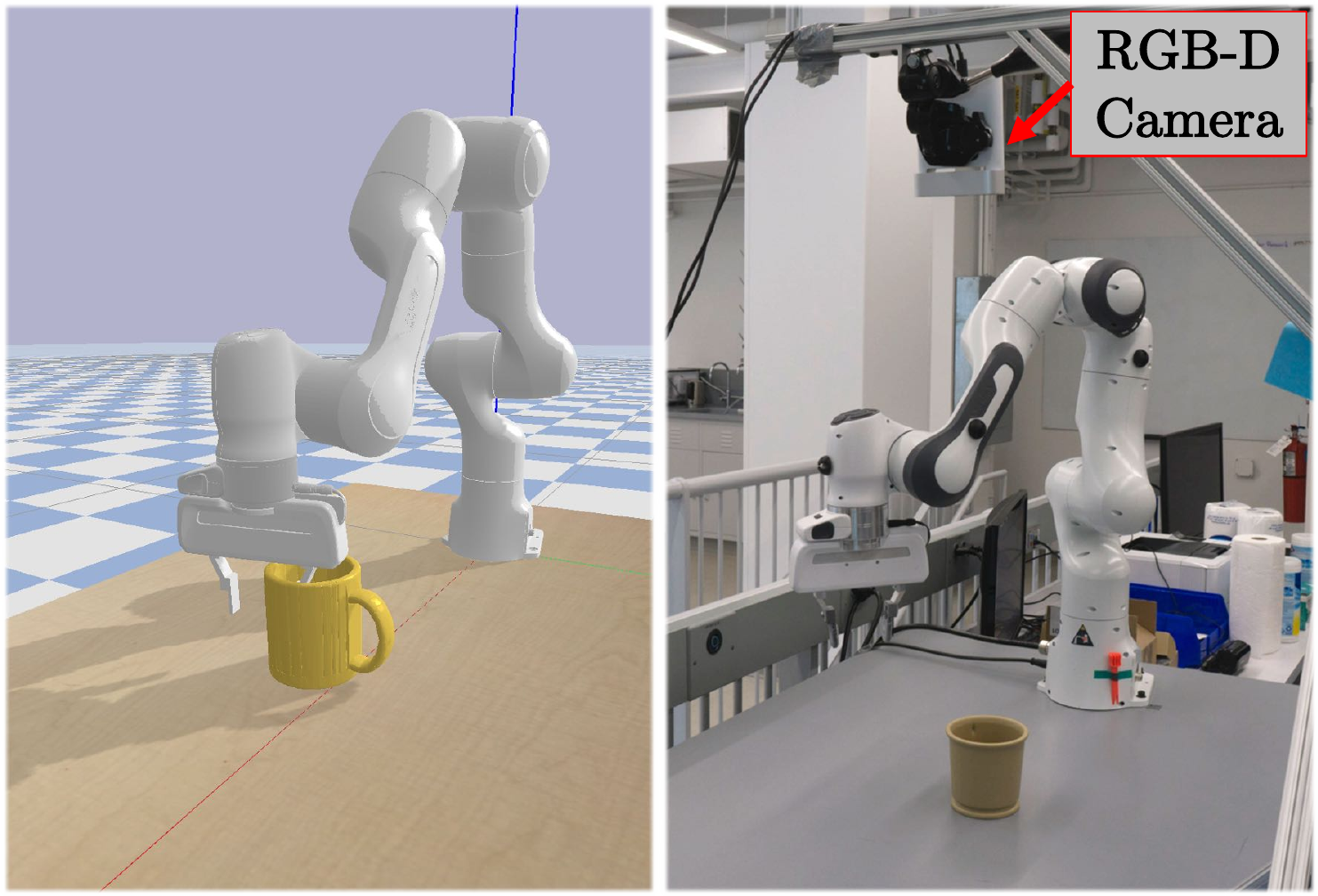}
\end{center}
\vspace{-10pt}
\caption{Grasping setup (left) in PyBullet simulation and (right) on hardware.}
\label{fig:grasp-setup}
\end{wrapfigure}

\vspace{-5pt}
The goal is to grasp and lift a mug from the table using a Franka Panda Arm (Fig.~\ref{fig:grasp-setup}). An open-loop action $a_{\textup{grasp}} = (x,y,z,\theta)$ is applied for each rollout and corresponds to desired 3D positions and yaw orientation of the grasp. The action is computed based on an observation of a $128 \times 128$ depth image from an overhead camera. 

We gathered 50 mugs of diverse shapes and dimensions from the ShapeNet dataset \citep{chang2015shapenet}. These mugs are split into 3 sets for expert demonstrations, PAC-Bayes Control training environments $S$, and test environments $S'$. They are then randomly scaled in all dimensions into multiple different mugs. Their masses are sampled from a uniform distribution. Each training or test environment consists of a unique mug from the set and a unique initial SE(2) pose on the table. A rollout is considered successful (zero cost) if the center of mass (COM) of the mug is lifted by 10 cm and the gripper palm makes no contact with the mug; otherwise, a cost of 1 is assigned to the rollout.

\begin{figure}[t]
  \begin{minipage}[c]{0.65\textwidth}
    \includegraphics[width=\textwidth]{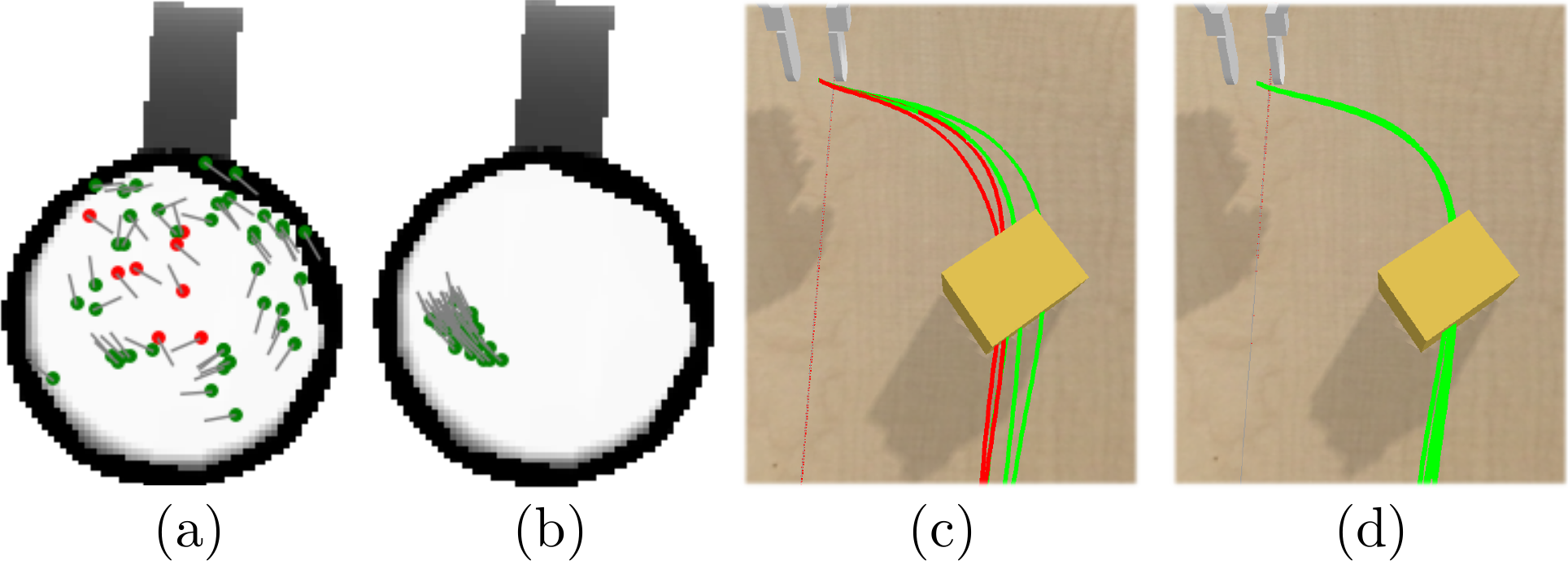}
  \end{minipage}\hfill
  \begin{minipage}[c]{0.32\textwidth}
    \caption{Sampled grasps [(a),(b)] and gripper trajectories when pushing boxes [(c),(d)] before and after polices being ``fine-tuned''. Grey tails in (a), (b) indicate grasp orientation. Red indicates failure and green indicates success. PAC-Bayes Control training ``shrinks'' the space of actions applied by policies.}
    \label{fig:grasp-push-samples}
  \end{minipage}
\vspace{-20pt}
\end{figure}

\textbf{Expert data.} In each of 60 environments, we specify 5 grasp poses along the rim. The initial depth image of the scene and corresponding grasp poses are recorded. It takes about an hour to collect the 300 trials.  

\textbf{Prior performance.} Each pair of initial observation and action from expert data is embedded into a latent $z \in \mathbb{R}^{10}$. Thus the length of time sequence in each demonstration $\zeta_i$ is $T_\textup{grasp}=1$. The reconstruction loss $\mathcal{L}_\textup{rec,grasp}$ is a combination of $l_2$ and $l_1$ loss between predicted and expert's actions. The prior policy distribution achieves 83.3\% success in novel environments in simulation. Shown in Fig.~\ref{fig:grasp-push-samples}(a), the prior $P_0$ captures the multi-modality of expert data: different latent $z$ sampled from $P_0$ generate a diverse set of grasps along the rim.

\textbf{Posterior performance.} $N=500$ environments are used for fine-tuning via PAC-Bayes. The resulting PAC-Bayes
bound $C_{\textup{bound}}^*:=C_{\textup{bound}}(P)$ of the posterior $P$ is 0.070. Thus, with probability 0.99 the optimized posterior policy is guaranteed to have an expected success rate of 93.0\% in novel environments (assuming that they are drawn from the same underlying distribution $\mathcal{D}$ as the training examples). The policy is then evaluated on 500 test environments in simulation and the success rate is 98.4\%. Fig.~\ref{fig:grasp-push-samples}(b) shows sampled grasps of a mug using the posterior distribution, which are concentrated at a relatively fixed position compared to grasps along the rim sampled from the prior.

\begin{table}[h!]
\footnotesize
\begin{center}
  \begin{tabular}{| c | c | c | c |}
    \hline
    \multirow{2}{*}{$1-C_{\textup{bound}}^*$} &
    \multicolumn{3}{c|}{True expected success (estimate)} \rule{0pt}{2ex}\\
    \cline{2-4}
     & Prior in simulation \rule{0pt}{2ex} & Posterior in simulation & Posterior on hardware \\ 
    \hline \hline
    0.930 \rule{0pt}{2ex} & 0.833 & 0.984 & 1.000, 1.000, 0.960 \\ \hline
  \end{tabular}
\vspace{5pt}
\caption{Prior performance, posterior performance, and generalization guarantees in grasping mug example using $N=500$ training environments.} 
\label{tab:grasping}
\end{center}
\end{table}

\textbf{Hardware implementation.} The posterior policy distribution trained in simulation is deployed on the hardware setup (Fig.~\ref{fig:grasp-setup}) in a zero-shot manner. We pick 25 mugs with a wide variety of shapes and materials (Fig.~\ref{fig:mugs}). Among three sets of experiments with different seeds (for sampling initial poses and latent $z$), the success rates are 100\% (25/25), 100\% (25/25), and 96\% (24/25). The hardware results thus validate the PAC-Bayes bound trained in simulation (Table~\ref{tab:grasping}).

\vspace{-5pt}
\subsection{Planar box pushing with real-time visual feedback}
\label{subsec:push}
\vspace{-5pt}

In this example, we tackle the challenging task of pushing boxes of a wide range of dimensions to a target region across the table. Real-time external visual feedback is applied using a $150 \times 150$ overhead depth image of the whole environment at 5Hz. The observation comprises of the depth map augmented with the proprioceptive x/y positions of the end-effector. The action is the desired relative-to-current x/y displacement of the end-effector $a_{\textup{push}}=(\Delta x, \Delta y)$. A low-level Jacobian-based controller tracks desired pose setpoints. The gripper fingers maintains a fixed height from the table and a fixed orientation, and the gripper width is set to 3 cm to maintain a two-point contact.

\begin{wrapfigure}[12]{r}{0.30\textwidth}
\vspace{-18pt}
\begin{center}
\includegraphics[width=0.30\textwidth]{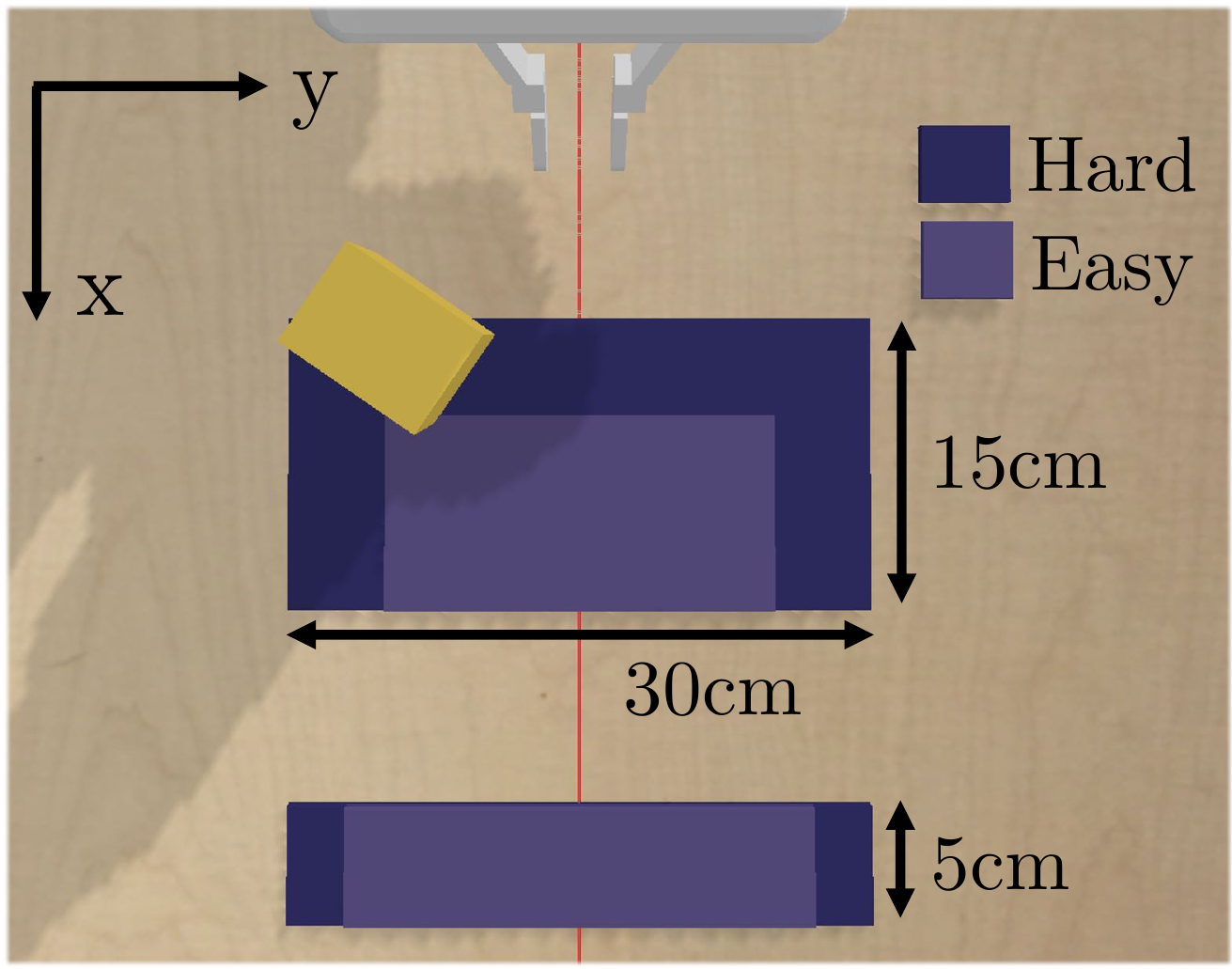}
\end{center}
\vspace{-10pt}
\caption{Starting and target regions for ``Easy'' and ``Hard'' tasks (``Hard'' overlaps ``Easy'').}
\label{fig:pushing-region}
\end{wrapfigure}

Rectangular boxes are generated by sampling the three dimensions (4-8 cm in length, 6-10 cm in width, and 5-8 cm in height) and the mass (0.1-0.2 kg). Each environment again consists of a unique box and a unique initial SE(2) pose. Based on the dimensions of starting and target regions, we define an ``Easy'' task and a ``Hard'' one (Fig.~\ref{fig:pushing-region}). A continuous cost is assigned based on how far the COM of the box is from the target region at the end of a rollout. 

\textbf{Expert data.} In each of 150 environments, we specify 2 different, successful pushing trajectories. The overhead depth image, the end-effector pose, and the desired end-effector displacement are recorded at 5Hz. It takes about 90 minutes to collect the 300 trials. 

\textbf{Prior performance.} Entire trajectories of observations and actions from expert data are embedded into latent $z \in \mathbb{R}^{5}$. Thus $T_\textup{push}$ is the total number of steps in each trial. The reconstruction loss $\mathcal{L}_\textup{rec,push}$ is again a combination of $l_2$ and $l_1$ loss between predicted and expert's actions. The prior policy distribution is able to achieve 84.2\% success in novel environments in the ``Easy'' task, and 74.9\% in the ``Hard'' task. Fig.~\ref{fig:grasp-push-samples}(c) shows a challenging environment of ``Hard'' task where the box starts far away from the red centerline and with a large yaw angle relative to the gripper. While experts always push at the corner of the box in demonstrations, the prior $P_0$ learned via behavioral cloning fails to imitate this behavior perfectly.

\textbf{Posterior performance.} We train both tasks using 500, 1000, and 2000 training environments. The resulting PAC-Bayes bound and empirical success rates across 2000 test environments are shown in Table \ref{tab:pushing}. Posterior performances are improved by about $10\%$ from the priors. Fig.~\ref{fig:grasp-push-samples}(c,d) shows that compared to prior policies, posterior policies perform better at the challenging task as the robot learns to push at the corner consistently. 

\begin{table}[H]
\footnotesize
\vspace{-8pt}
\begin{center}
\begin{tabular}{| c | c | c | c | c | c |} \hline
\multirow{3}{*}{\makecell{Task \\ difficulty}} &
\multirow{3}{*}{\makecell{$N$ ($\#$ of training \\ environments)}} &
\multirow{3}{*}{$1-C_{\textup{bound}}^*$} &
\multicolumn{3}{c|}{True expected success (estimate)} \rule{0pt}{2ex}\\ 
\cline{4-6} 
 &  &  & \makecell{Prior in \rule{0pt}{2ex} \\ simulation } & \makecell{Posterior \rule{0pt}{2ex} in \\ simulation} & \makecell{Posterior \rule{0pt}{2ex} on \\ hardware} \\ 
\hline \hline
Easy \rule{0pt}{2ex}& 500 & 0.861 & \multirow{3}{*}{0.842} & 0.929 &  - \\
\cline{1-3} \cline{5-6} 
Easy \rule{0pt}{2ex}& 1000 & 0.888 & & 0.937 & 0.800, 0.867, 0.933 \\ 
\cline{1-3} \cline{5-6} 
Easy \rule{0pt}{2ex}& 2000 & 0.904 & & 0.945 & - \\ 
\hline \hline
Hard \rule{0pt}{2ex}& 500 & 0.754 & \multirow{3}{*}{0.749} & 0.863 &  - \\
\cline{1-3} \cline{5-6} 
Hard \rule{0pt}{2ex}& 1000 & 0.791 & & 0.864 & 0.800, 0.800, 0.800 \\
\cline{1-3} \cline{5-6} 
Hard \rule{0pt}{2ex}& 2000 & 0.810 & & 0.864 & - \\
\hline
\end{tabular}
\vspace{5pt}
\caption{Prior performance, posterior performance, and generalization guarantees in pushing box example.}
\label{tab:pushing}
\end{center}
\vspace{-25pt}
\end{table}

\textbf{Hardware implementation.} The posterior policy distributions trained using 1000 environments are deployed on the real arm. For both ``Easy'' and ``Hard'' tasks, three set of experiments with different seeds are performed with 15 rectangular blocks (Fig.~\ref{fig:boxes}). The success rates are shown in Table \ref{tab:pushing}. Note that the result for the ``Easy'' task falls short of the bound in some seeds. We suspect that the sim2real performance is affected by imperfect depth images from the real camera and minor differences in dynamics between simulation and the actual arm.

\vspace{-5pt}
\subsection{Vision-based indoor navigation}
\label{subsec:navigate}
\begin{figure}[t]
\centering
\includegraphics[width=0.85\textwidth]{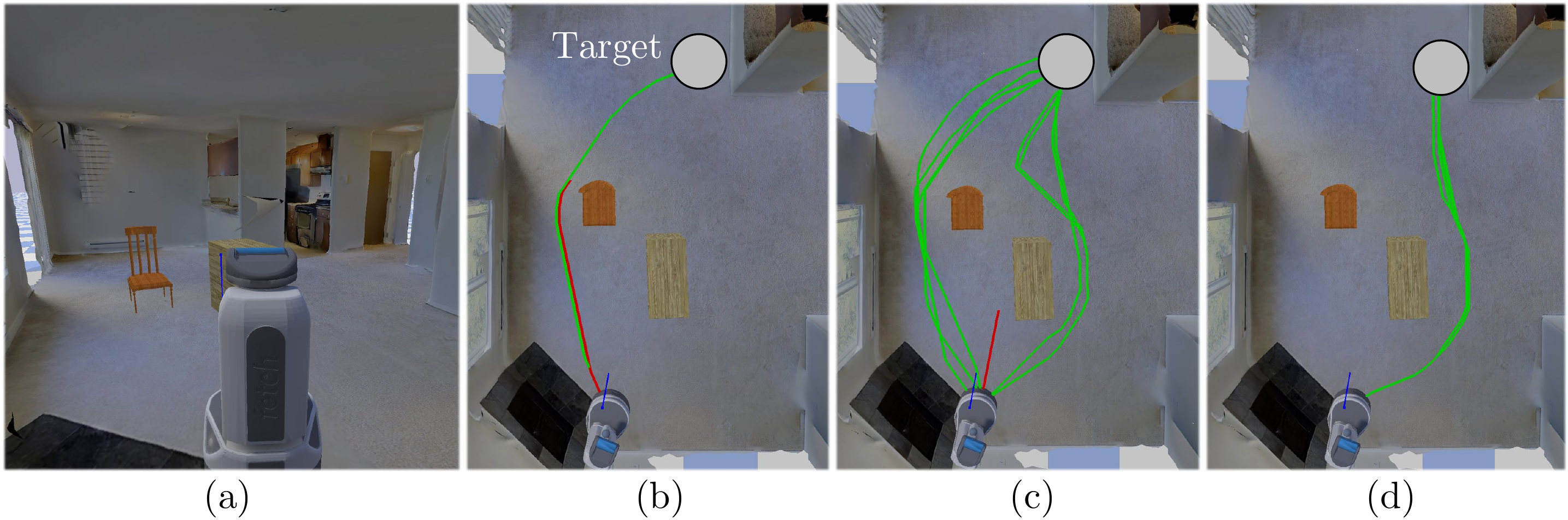}
\vspace{-5pt}
\caption{(a) robot view in navigation, (b) 10 trajectories sampled from a single-modal prior, (c) from a multi-modal prior, and (d) from a posterior. Red indicates failure and green indicates success.}
\label{fig:nav-samples}
\vspace{-10pt}
\end{figure}

\vspace{-5pt}
In this example, a Fetch mobile robot needs to navigate around furniture of different shapes, sizes, and initial poses, before reaching a target region in a home environment (Fig.~\ref{fig:nav-samples}). We use iGibson \citep{xia2020interactive} to render photorealistic visual feedback in PyBullet simulations. Again we consider a challenging setting where the robot executes actions given front-view camera images and without any extra knowledge of the map. At each step, the policy takes a $200 \times 200$ RGB-D image and chooses the motion primitive with the highest probability from the four choices (move forward, move backward, turn left, and turn right), $a_{\textup{nav}}= \operatorname*{arg\,max}(p_{\textup{forward}}, p_{\textup{backward}}, p_{\textup{left}}, p_{\textup{right}})$.

We collect 293 tables and 266 chairs from the ShapeNet dataset \citep{chang2015shapenet}. A fixed scene (Sodaville) from iGibson  is used for all environments. One table and one chair are randomly spawned between the fixed starting location and the target region in each environment. The SE(2) poses of the furniture are drawn from a uniform distribution. A rollout is successful if the robot reaches the target within 100 steps without colliding with the furniture and the wall.

\textbf{Expert data.} In each of 100 environments, two different, successful robot trajectories are collected. The front-view RGB-D image from the robot and the motion primitive applied at each step are recorded. It takes about 90 minutes to collect the 200 demonstrations.  

\textbf{Prior performance.}
Short sequences ($T_\textup{nav}=3$) of observations and actions from expert data are embedded into latent $z \in \mathbb{R}^{10}$. The reconstruction loss $\mathcal{L}_\textup{rec,nav}$ is the cross-entropy loss between predicted action probabilities and expert's actions. Before each rollout, a single latent $z$ is sampled and then applied as the policy for all steps. The prior policy distribution is able to achieve 65.4\% success in novel environments in simulation. Fig.~\ref{fig:nav-samples}(c) shows the diverse trajectories generated by the trained cVAE.

\textbf{Posterior performance.} The prior is ``fine-tuned'' using 500 and 1000 training environments. The resulting PAC-Bayes bound and empirical success rates across 2000 test environments are shown in Table \ref{tab:navigation}. Fig.~\ref{fig:nav-samples}(d) shows that similar to grasping and pushing examples, the robot follows relatively the same trajectories in the same environment when executing ``fine-tuned'' posterior policies. In that environment, the robot now consistently chooses the bigger gap on the right to navigate and avoids the narrow one on the left chosen by some prior policies.
\begin{table}[h!]
\footnotesize
\vspace{-5pt}
\begin{center}
\begin{tabular}{| c | c | c | c |} \hline
\multirow{3}{*}{\makecell{$N$ ($\#$ of training \\ environments)}} &
\multirow{3}{*}{$1-C_{\textup{bound}}^*$} &
\multicolumn{2}{c|}{True expected success (estimate)} \rule{0pt}{2ex}\\ 
\cline{3-4} 
 &  & \makecell{Prior in \rule{0pt}{2ex}\\ simulation} & \makecell{Posterior in \rule{0pt}{2ex}\\ simulation} \\ 
\hline \hline
500 \rule{0pt}{2ex} & 0.723 & \multirow{2}{*}{0.654} & 0.791 \\
\cline{1-2} \cline{4-4}
1000 \rule{0pt}{2ex} & 0.741 & & 0.799  \\ 
\hline
\end{tabular}
\vspace{5pt}
\caption{Prior performance, posterior performance, and generalization guarantees in indoor navigation example}
\label{tab:navigation}
\end{center}
\vspace{-25pt}
\end{table}

We also investigate the benefit of ``fine-tuning'' a multi-modal prior distribution vs. a single-modal one. Fig.~\ref{fig:nav-samples}(b,c) show the differences in the policies sampled from these two priors in the same environment. We find that the multi-modality of the prior accelerates PAC-Bayes training and leads to better empirical performance of the posterior. Please refer to \ref{sup:benefit} for full discussions.

\vspace{-5pt}
\section{Conclusion}
\label{sec:conclusion}
\vspace{-5pt}
We have presented a framework for providing generalization guarantees for policies learned via imitation learning. Policies are trained through two stages: (i) a “prior” policy distribution is learned through multi-modal behavior cloning to mimic an expert's behavior, and (ii) the prior is then ``fine-tuned'' using PAC-Bayes Control framework by explicitly optimizing the generalization bound. The resulting ``posterior'' distribution $P$ over policies achieves strong empirical performance and generalization guarantees in novel environments, which are verified in simulation of manipulation and navigation tasks, as well in hardware experiments of manipulation tasks.

\emph{Challenges and future work:} we find training the cVAE requires significant tuning and can be difficult with embedding long sequences of high-dimensional images as input. This limits the framework from handling long-horizon tasks where longer sequences (if not whole trajectories) of observation/action pairs need to be embedded in the latent space and information from images varies significantly along the rollout. For example, tasks like pouring water involve multiple stages of manipulation --- picking up the mug, rotating it to pour, and putting it back on the table. We are looking into learning separate latent distributions that encode different stages of the task. The other exciting direction is to apply the Dataset Augmentation (``DAgger'') Algorithm \citep{ross2011reduction} that constantly injects additional expert knowledge into training as the policy is refined.


\acknowledgments{The authors were partially supported by the Office of Naval Research [Award Number: N00014-18-1-2873], the Google Faculty Research Award, and the Amazon Research Award.}


\bibliography{bib-pac-imitation}


\renewcommand{\thetable}{A\arabic{table}}
\renewcommand{\theequation}{A\arabic{equation}}
\renewcommand\thefigure{A\arabic{figure}}
\renewcommand{\thesubsection}{A\arabic{subsection}}
\setcounter{figure}{0}
\setcounter{table}{0}
\setcounter{equation}{0}

\clearpage
\section*{Appendix}
\label{sec:appendix}
\subsection{Natural Evolutionary Strategies}
\vspace{-5pt}
\label{sup:es}
Below are the implementation details for training the posterior distribution with Natural Evolutionary Strategies (NES) (Sec.~\ref{sec:pac-es}). Our objective is to minimize the upper bound $C_\PAC$ in Thm.~\ref{thm:pac bayes control}. We will first express the gradient of $C_\PAC$ w.r.t. $\psi:=(\mu,\sigma)$ as follows (using \eqref{eq:emp-ES-grad} and \eqref{eq:reg-ES-grad}):
\small
\begin{align}
    \nabla_\psi C_{\PAC}(P, P_0) & = \nabla_\psi C_S(P) + \nabla_\psi \sqrt{R(P, P_0)} \enspace, \nonumber\\
    & = \nabla_\psi C_S(P) + \frac{1}{2 \sqrt{R}}\nabla_\psi R(P, P_0) \enspace, \nonumber\\
    & = \frac{1}{N} \sum_{E \in S}\mathop{\mathbb{E}}_{z \sim \normal_\psi}[C(\pi_{\scaleto{\theta^*}{4pt},z};E)\nabla_\psi \ln \normal_{\psi}(z)] + \frac{1}{4N\sqrt{R}}\nabla_\psi \mathop{\mathbb{E}}_{z \sim \normal_\psi}\Big[\log\frac{\normal_\psi(z)}{\normal_{\psi_0}(z)}\Big] \nonumber \\
    & = \frac{1}{N} \sum_{E \in S}\mathop{\mathbb{E}}_{z \sim \normal_\psi}[C(\pi_{\scaleto{\theta^*}{4pt},z};E)\nabla_\psi \ln \normal_{\psi}(z)] + \frac{1}{4N\sqrt{R}} \mathop{\mathbb{E}}_{z \sim \normal_\psi}\Big[\log\frac{\normal_\psi(z)}{\normal_{\psi_0}(z)} \nabla_\psi \ln \normal_{\psi}(z)\Big] \nonumber \nonumber\\
    & = \frac{1}{N} \sum_{E \in S}\mathop{\mathbb{E}}_{z \sim \normal_\psi}\Big[ \underbrace{\Big(C(\pi_{\scaleto{\theta^*}{4pt},z};E) + \frac{1}{4N\sqrt{R}} \log\frac{\normal_\psi(z)}{\normal_{\psi_0}(z)}\Big) \nabla_\psi \ln \normal_{\psi}(z)}_{C_{ES}(z;E)}\Big] \nonumber
\end{align}
\normalsize
To estimate the expectation on the right-hand side (RHS) of the above equation, we sample a few $z$'s from $\mathcal{N}_\psi$ and average $C_{ES}$ over them. In particular, we perform antithetic sampling of $z$ (i.e., for each $z$ that we sample, we also evaluate $C_{ES}$ for $-z$) to reduce the variance of the gradient estimate \citep{salimans2017evolution}. This gives us the following ES gradient estimate for $m$ samples of $z$ ($2m$ with antithetic sampling):
\begin{equation}\nonumber
    \nabla_\psi C_{\PAC}(P, P_0) \approx \frac{1}{N} \sum_{E \in S} \bigg[\frac{1}{2m}\sum_{i=1}^{m}(C_{ES}(z_i;E)+C_{ES}(-z_i;E))\bigg]
\end{equation}
Finally, using the Fisher information matrix $F_\psi$ (of the normal distribution $\mathcal{N}_\psi$ w.r.t. $\psi$) we compute an estimate of the natural gradient \cite{wierstra2014natural}, denoted by $\tilde{\nabla}_\psi$, from the above equation:
\begin{equation}
    \tilde{\nabla}_\psi C_{\PAC}(P, P_0) \approx F_\psi^{-1} \frac{1}{N} \sum_{E \in S} \bigg[\frac{1}{2m}\sum_{i=1}^{m}(C_{ES}(z_i;E)+C_{ES}(-z_i;E))\bigg] . \nonumber
\end{equation}
The natural gradient estimate computed above is passed to the Adam optimizer \cite{kingma2014adam} to update the belief distribution parameterized by $\psi$. In practice we use $\psi=(\mu,\log \sigma^2)$ instead of $(\mu,\sigma)$ to avoid imposing the strict positivity constraint on $\sigma$ for the gradient update with Adam.

\subsection{Derivations of the final bound}
\vspace{-5pt}
\label{sup:final-bound}
The derivations follow \citep{Majumdar19, Veer20}. Following Sec.~\ref{sec:pac-es}, first the empirical training cost is estimated by sampling a large number of policies $z_1, ..., z_L$ from the optimized posterior distribution $\normal_{\psi^*}$, and averaging over all $N$ training environments in $S$:
\begin{equation}
    \hat{C}_S(\normal_\psi^*) := \frac{1}{NL}\sum_{E \in S} \sum_{i=1}^L C(\pi_{\scaleto{\theta^*}{4pt},z_i}; E).\label{eq-app:sample-cost}
\end{equation}
Next, the error between $\hat{C}_S(\normal_{\psi^*})$ and $C_S(\normal_{\psi^*})$ can be bounded using a sample convergence bound \citep{langford2002not} $\bar{C}_S$, which is an application of the relative entropy verision of the Chernoff bound for random variables (i.e., costs) bounded in $[0,1]$ and holds with probability $1-\delta'$:
\begin{equation}
    C_S(\normal_{\psi^*}) \leq \bar{C}_S(\normal_{\psi^*};L,\delta') := \KL^{-1}(\hat{C}_S(\normal_{\psi^*}) || \frac{1}{L} \log (\frac{2}{\delta'})).\label{eq-app:sample-convergence-bound}
\end{equation}
where $\KL^{-1}$ refers to the KL-inverse function and can be computed using a Relative Entropy Program (REP) \citep{Majumdar19}. $\KL^{-1}: [0,1] \times [0,\infty) \rightarrow [0,1]$ is defined as:
\begin{equation}
\KL^{-1}(p||c) := \sup \{q \in [0,1] \ | \ \KL(p||q) \leq c \}.\label{eq-app:kl-inv}
\end{equation}
The KL-inverse function can also provide the following bound on the true expected cost $C_{\mathcal{D}}(P)$ (Theorem 2 from \citep{Veer20}):
\begin{equation}
    C_{\mathcal{D}}(P) \leq \KL^{-1} \Big (C_S(P) || \frac{\KL(P||P_0) + \log \frac{2 \sqrt{N}}{\delta}}{N} \Big ).\label{eq-app:kl-inv-bound}
\end{equation}
Now combining inequalities \eqref{eq-app:sample-convergence-bound} and \eqref{eq-app:kl-inv-bound} using the union bound, the following final bound $C_\textup{bound}$ holds with probability at least $1-\delta-\delta'$:
\begin{equation}
    C_\mathcal{D}(\normal_{\psi^*}) \leq C_\textup{bound}(\normal_{\psi^*}) := \KL^{-1} \bigg (\bar{C}_S(\normal_{\psi^*};L,\delta') || \frac{\KL(\normal_{\psi^*} || P_0) + \log \frac{2 \sqrt{N}}{\delta}}{N} \bigg ).\label{eq-app:final-bound}
\end{equation}

\subsection{Algorithms for the two training stages}
\vspace{-5pt}
\label{sup:full-algorithm}
\begin{algorithm}[H]
\caption{Multi-modal Behavioral Cloning} 
\begin{algorithmic}[1]
    \Require $\resizebox{33pt}{!}{$q_\phi, \pi_{\scaleto{\theta}{4pt},z}$}, \{\zeta_i\}_{i=1}^M$ \Comment{Encoder network, decoder network, demonstrations}
	\For{$i=1,\cdots,n_\textup{iter}$}
		\State $\phi, \theta \leftarrow \arg \min_{\phi, \theta} \sum_{\zeta_i} [(\sum_{t=1}^T \mathcal{L}_{\rm rec}(\pi_{\scaleto{\theta}{4pt},z}(o_t),a_t))-\lambda D_{\KL}((q_{\phi}(z | o_{1:T}, a_{1:T})|| p(z))]$ \\ \Comment{cVAE training} 
	\EndFor \\
	\Return $\resizebox{22pt}{!}{$\pi_{\scaleto{\theta^*}{4pt},z}$}, P_0 = \normal_{\psi_0} := \normal(0,I)$ \Comment{Optimized decoder network, prior over policies}
	\end{algorithmic}
\normalsize
\end{algorithm}
\vspace{-13pt}
\begin{algorithm}[H]
\caption{PAC-Bayes Policy ``Fine-tuning''}
\begin{algorithmic}[1]
    \Require $\resizebox{22pt}{!}{$\pi_{\scaleto{\theta^*}{4pt},z}$}, P_0, S=\{E_1,...,E_N\}, \delta,\delta'$ \Comment{Policy (decoder) network, prior over policies} \\ \Comment{training environments, probability thresholds}
	\For{$i=1,\cdots,n_\textup{iter}$}
		\State Sample $z_{j,k} \sim P$ and $-z_{j,k}$ for $j=1,...,N, k=1,...,m$ 
		\State Perform rollouts to get cost $C(\pi_{\scaleto{\theta^*}{4pt},z_{j,k}};E_j)$
		\State Compute $\tilde{\nabla}_\psi C_{\PAC}(P, P_0)$, and update $P = \normal_\psi$
	\EndFor
	\State Sample $z_{j,l} \sim P$ for $j=1,...N, l=1,...,L$ 
	\State Perform rollouts to get the estimate empirical training cost $\hat{C}_S(P)$ using \eqref{eq-app:sample-cost}
	\State Compute the sample convergence bound $\bar{C}_S(P)$ using \eqref{eq-app:sample-convergence-bound}
	\State Compute the final bound $C_\textup{bound}(P)$ using \eqref{eq-app:final-bound}\\
	\Return $P= \normal_{\psi^*}, C_\textup{bound}(P)$ \Comment{Posterior over policies, PAC-Bayes generalization bound}
	\end{algorithmic}
\normalsize
\end{algorithm}

\subsection{Benefit of using a multi-modal prior policy distribution}
\vspace{-5pt}
\label{sup:benefit}
\begin{wrapfigure}[14]{r}{0.35\textwidth}
\vspace{-20pt}
\begin{center}
\includegraphics[width=0.35\textwidth]{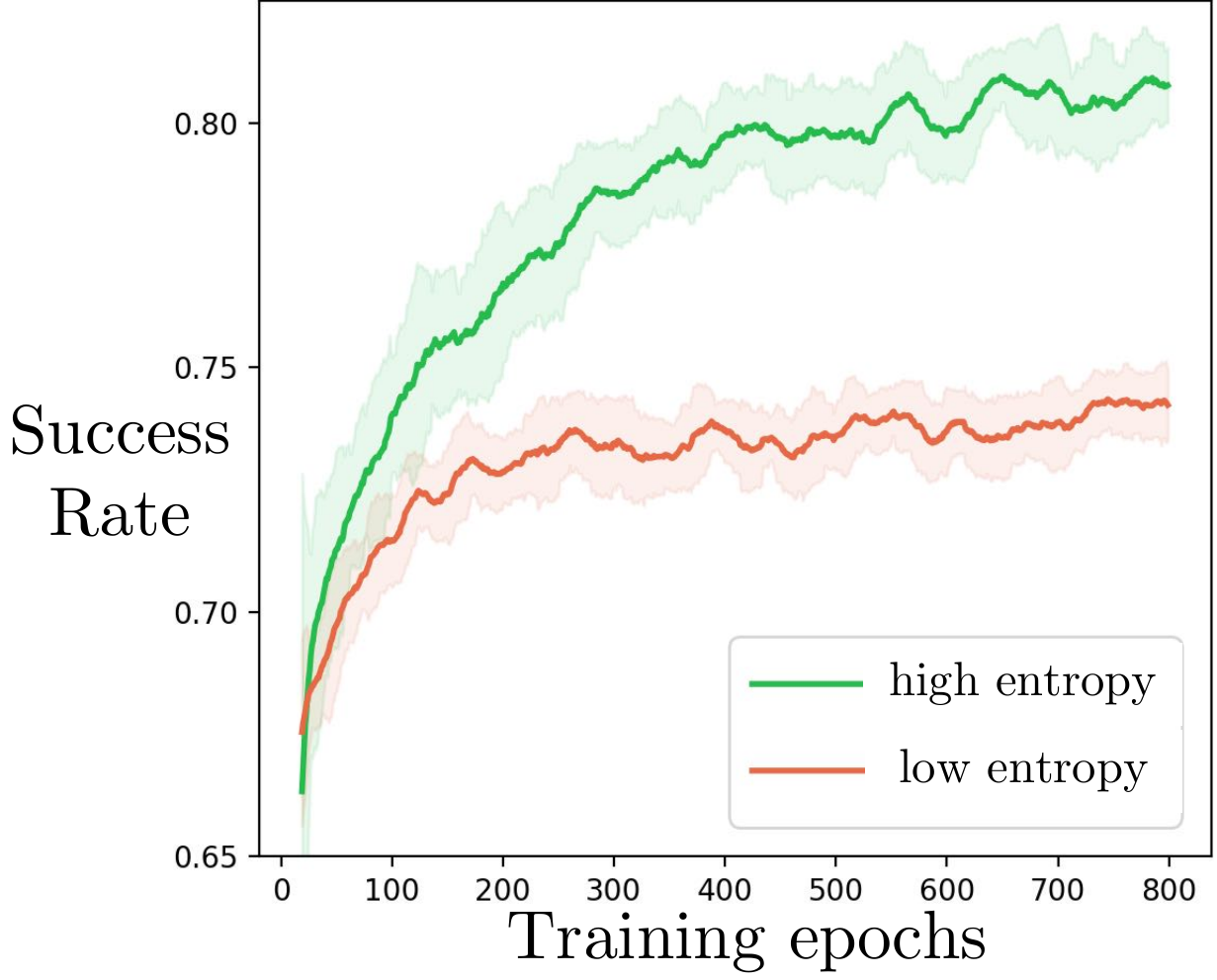}
\end{center}
\vspace{-10pt}
\caption{Empirical success rates over all training environments when ``fine-tuning'' priors with low entropy vs. higher entropy.}
\label{fig:kl-comparison}
\end{wrapfigure}
As shown in Fig.~\ref{fig:nav-samples}, the prior policy distribution for indoor navigation can exhibit either uni-modal or multi-modal behavior. Uni-modal prior has low entropy while multi-modal prior has higher entropy. We expect that a prior with high entropy can benefit PAC-Bayes ``fine-tuning'' as opposed to a low-entropy prior; we investigate this further in the indoor navigation example and report the training curves with the low- and high-entropy priors in Fig.~\ref{fig:kl-comparison}. While the two priors achieve similar empirical performance before being ``fine-tuned'', the prior with higher entropy trains faster and achieves better empirical performance on test environments at the end of ``fine-tuning'' (0.799 vs. 0.706 in success rate).

As illustrated in Fig.~\ref{fig:nav-samples}, the prior distribution with low entropy always picks the small gap on the left to navigate. Although the policies were successful in behavior cloning environments, they might fail in the ``fine-tuning'' training environments where the gap can shrink or there can be an occluded piece of furniture behind the gap. Hence ``fine-tuning'' can be difficult as the prior always chooses that specific route. Instead, the prior with higher entropy ``encourages'' the robot to try different directions, and is intuitively easier to ``adapt'' to new environments. ``Fine-tuning'' then picks the better policy among all available ones for each environment.

\vspace{-5pt}
\subsection{Synthetic training details}
\vspace{-5pt}
\label{sup:synthetic-training}
All behavioral cloning training is run on a desktop machine with Intel i9-9820X CPU and a Nvidia Titan RTX GPU. PAC-Bayes training for manipulation tasks is performed on an Amazon Web Services (AWS) c5.24xlarge instance that has 96 threads. PAC-Bayes training for the navigation task is done using an AWS g4dn.metal instance that has 64 threads and 8 Nvidia Tesla T4 GPUs. It took about 3 hours to fine-tune the grasping policy with 500 training environments, and about 8 hours for pushing with 500 environments (a GPU instance could have been used to accelerate model inferences). The navigation task is more computationally intensive as it requires GPUs to render the indoor scene - it took about 30 hours with 500 training environments.

\paragraph{Choice of latent dimensions} We find that a relatively small latent dimension (i.e. less than 10) is sufficient to encode multi-modality of the demonstrations. We use 10 for both grasping and indoor navigation, and 5 for pushing as the demonstrations are less multi-modal. It is possible to use an even smaller dimension and achieve similar empirical performance of the prior, but posterior performance could suffer as the small dimension constrains fine-tuning. We also find difficulty in learning a structured latent space of higher dimension (i.e. more than 20) for behavioral cloning.
\vspace{-5pt}
\subsubsection{Grasping mugs}
\textbf{cVAE architecture.}
\begin{figure}[H]
\vspace{-5pt}
\centering
\includegraphics[width=\textwidth]{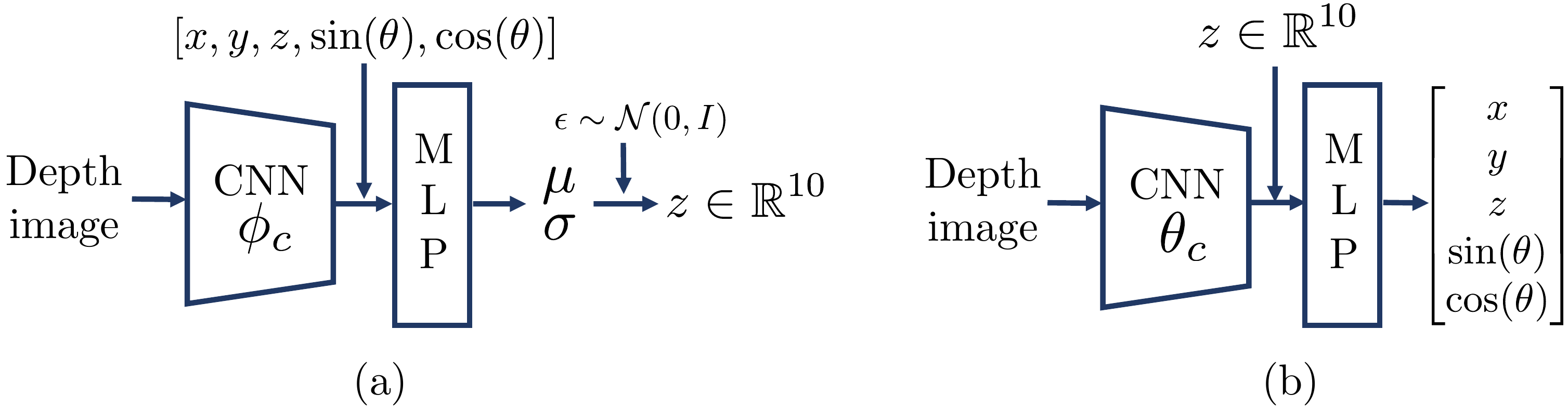}
\caption{(a) Encoder, and (b) decoder of the cVAE for grasping mugs example.} 
\label{fig:cvae-grasp}
\vspace{-15pt}
\end{figure}
In both the encoder and decoder, image features are generated through a spatial-softmax layer after the convolutional layers (CNN). In the encoder, the action is appended to image features before being passed into a multi-layer perceptron (MLP). In the decoder, the sampled latent $z$ is appended to the image features before being passed into an MLP. Sine and cosine encodings are used for yaw angle action. A learning rate of 1e-3 and a weight decay rate of 1e-5 are used for cVAE training.

\textbf{Reconstruction loss function of the CVAE.} $\mathcal{L}_\textup{rec,grasp}$ is a combination of $l_1$ and $l_2$ losses between the predicted actions and the expert's:
\begin{equation}\nonumber
    \mathcal{L}_\textup{rec,grasp} = ||a_\textup{pred} - a_\textup{expert}||_1 + 0.1 ||a_\textup{pred} - a_\textup{expert}||_2
\end{equation}
\textbf{Environment setup.}
\vspace{-5pt}
\begin{itemize}[leftmargin=0.3in]
    \setlength\itemsep{1pt}
    \item The diameter of the mugs is sampled uniformly from [8.5 cm, 13 cm].
    \item The mass of the mugs is sampled uniformly from [0.1 kg, 0.5 kg].
    \item The friction coefficients for the environment are 0.3 for lateral and 0.01 for torsional.
    \item The moment of inertia of the mugs is determined by the simulator assuming uniform density.
    \item The initial SE(2) pose of the mugs is sampled uniformly from [0.45 cm, 0.55 cm] in $x$, [-0.05 cm, 0.05 cm] in $y$, and [$-\pi$, $\pi$] in yaw (all relative to robot base).
\end{itemize}

\textbf{Posterior distribution (for $N=1000$ training environments).}

$\mu = [0.033, 0.319, 0.197, -0.846, -1.040, 0.052, -0.507, 0.273, -0.657, -0.357]$.

$\sigma = [1.062, 0.997, 0.923, 0.522, 0.229, 0.966, 1.012, 0.960,  0.236, 0.866]$.

\textbf{Final bound.} $C_\textup{bound}$ is computed using $\delta=0.009, \delta'=0.001$, and $L = 10000$.

\clearpage
\subsubsection{Pushing boxes}
\textbf{cVAE architecture.}
\begin{figure}[H]
\vspace{-5pt}
\centering
\includegraphics[width=\textwidth]{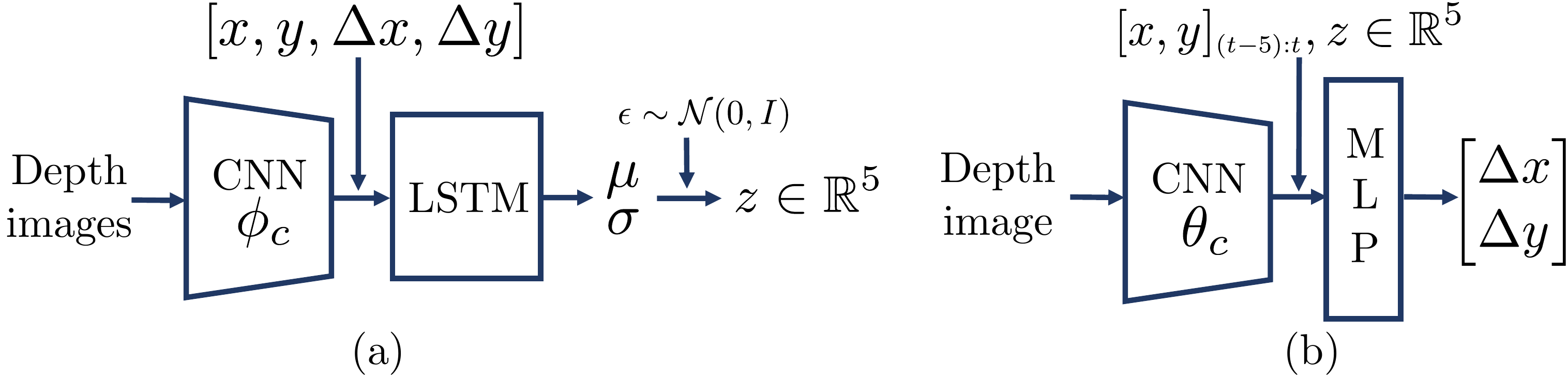}
\caption{(a) Encoder, and (b) decoder of the cVAE for pushing boxes example. }
\label{fig:cvae-push}
\vspace{-15pt}
\end{figure}
In both the encoder and decoder, image features are generated through a spatial-softmax layer after the CNN. In the encoder, the action ($\Delta x, \Delta y$) and the proprioceptive state ($x,y$) are appended to image features before being passed into an MLP. In the decoder, the sampled latent $z$ and a history (5 steps) of the proprioceptive state are appended to the image features before being passed into a MLP. A learning rate of 1e-3 and a weight decay rate of 1e-5 are used for cVAE training.

\textbf{Reconstruction loss function of the CVAE.} $\mathcal{L}_\textup{rec,push}$ is again a combination of $l_1$ and $l_2$ losses between the predicted actions and the expert's.
\begin{equation}\nonumber
    \mathcal{L}_\textup{rec,push} = ||a_\textup{pred} - a_\textup{expert}||_1 + 3||a_\textup{pred} - a_\textup{expert}||_2
\end{equation}
\textbf{Environment setup.}
\vspace{-5pt}
\begin{itemize}[leftmargin=0.3in]
    \setlength\itemsep{1pt}
    \item The dimensions of the rectangular boxes are sampled uniformly from [4 cm, 8 cm] in length, [6 cm, 10 cm] in width, and [5 cm, 8 cm] in height.
    \item The mass of the boxes is sampled uniformly from [0.1 kg, 0.2 kg].
    \item The friction coefficients for the environment are 0.3 for lateral and 0.01 for torsional.
    \item The moment of inertia of the boxes is determined by the simulator assuming uniform density.
    \item For the ``Easy'' task, the initial SE(2) pose of the boxes is sampled uniformly from [0.55 cm, 0.65 cm] in $x$, [-0.10 cm, 0.10 cm] in $y$, and [$-\pi/4$, $\pi/4$] in yaw (all relative to robot base). The dimensions of the target region are [0.75 cm, 0.80 cm] in $x$ and [-0.12 cm, 0.12 cm] in $y$.
    \item For the ``Hard'' task, the initial SE(2) pose of the boxes is sampled uniformly from [0.50 cm, 0.65 cm] in $x$, [-0.15 cm, 0.15 cm] in $y$, and [$-\pi/4$, $\pi/4$] in yaw (all relative to robot base). The dimensions of the target region are [0.75 cm, 0.80 cm] in $x$ and [-0.15 cm, 0.15 cm] in $y$.
\end{itemize}

\textbf{Posterior distribution (for $N=1000$ training environments).}

For the ``Easy'' task:

$\mu = [0.690, -0.098, -0.733, 0.361, 1.947]$, and $\sigma = [0.829, 0.373, 0.346, 0.281, 0.255]$.

For the ``Hard'' task:

$\mu = [0.868, 0.531, -0.877, 0.172, 2.114]$, and $\sigma = [0.892, 0.517, 0.491, 0.445, 0.421]$.

\textbf{Final bound.} $C_\textup{bound}$ is computed using $\delta=0.009, \delta'=0.001$, and $L = 25000$.

\clearpage
\subsubsection{Indoor navigation}
\textbf{cVAE architecture.}
\begin{figure}[H]
\vspace{-5pt}
\centering
\includegraphics[width=\textwidth]{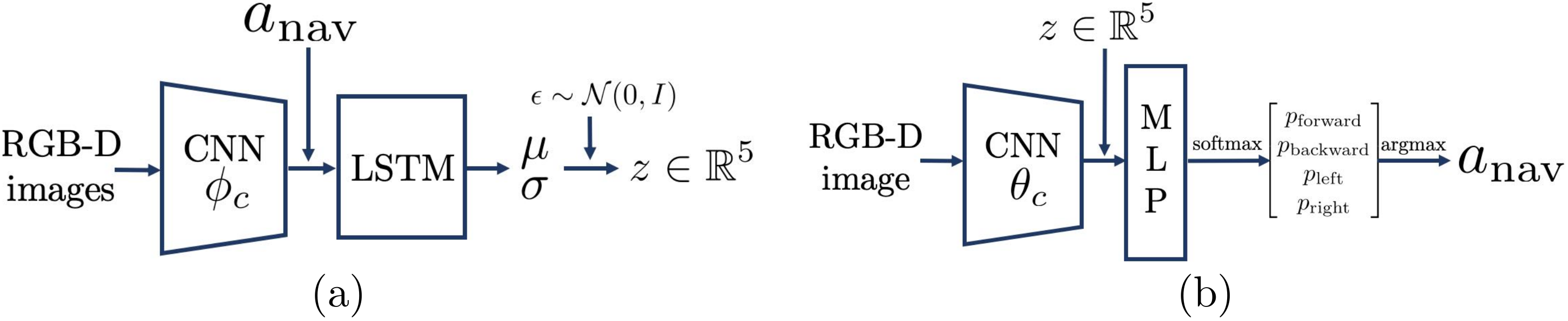}
\caption{(a) Encoder, and (b) decoder of the cVAE for indoor navigation example. }
\label{fig:cvae-nav}
\vspace{-15pt}
\end{figure}
The CNNs in the encoder and decoder ($\phi_c$ and $\theta_c$) share weights. Two separate CNNs are used for RGB and depth channels of the images. In both the encoder and decoder, image features are generated through a spatial-softmax layer after the CNN. In the encoder, the action (one-hot encoding of the motion primitives) is appended to image features before being passed into a single LSTM layer. In the decoder, the sampled latent $z$ is appended to the image features before being passed into an MLP. The output of the MLP is passed through a softmax layer to get normalized probabilities for the four motion primitives. Finally the action is chosen as the $\operatorname*{arg\,max}$ of the four. A learning rate of 1e-3 and a weight decay rate of 1e-5 are used for cVAE training.

\textbf{Reconstruction loss function of the CVAE.} $\mathcal{L}_\textup{rec,nav}$ is the cross entropy loss between the predicted action probabilities and the expert's. One-hot encoding is used for expert's actions.

\textbf{Environment setup.}
\vspace{-5pt}
\begin{itemize}[leftmargin=0.3in]
    \setlength\itemsep{1pt}
    \item The step length for ``move forward/backward'' is fixed as 20 cm, and the turning angle for ``turn left/right'' is fixed as 0.20 radians.
    \item Instead of physically simulating the robot movement, we change the base position and orientation of the robot at each step and check its collisions with the furniture and the wall.
    \item The arm on the Fetch robot is removed from the robot URDF file to save computations.
    \item The initial SE(2) pose of the furniture is sampled uniformly from [-3.0 m, -1.0 m] in $x$, [-1.0 m, 1.5 m] in $y$, and [$-\pi/2$, $\pi/2$] in yaw (all relative to the world origin in the Sodaville scene from iGibson \citep{xia2020interactive}).
    \item A red snack box from the YCB object dataset \citep{calli2015ycb} is placed at the target region in each environment.
\end{itemize}

\textbf{Posterior distribution (for $N=1000$ training environments).}

$\mu = [-1.600, 0.704, 1.330, 0.233, 0.004, 0.735, 1.447, 0.630, 1.843, 1.132]$.

$\sigma = [0.166, 0.885, 0.947, 0.973, 0.830, 0.816, 0.962, 0.860, 0.848, 0.965]$.

\textbf{Final bound.} The final bound $C_\textup{bound}$ is computed using $\delta=0.009, \delta'=0.001$, and $L = 25000$.

\subsection{Hardware experiment details}
\label{sup:hardware-experiments}
All hardware experiments are performed using a Franka Panda arm and a Microsoft Azure Kinect RGB-D camera. Robot Operating System (ROS) Melodic package (on Ubuntu 18.04) is used to integrate robot arm control and perception.
\subsubsection{Grasping mugs}
\begin{figure}[H]
\centering
\includegraphics[width=0.5\textwidth]{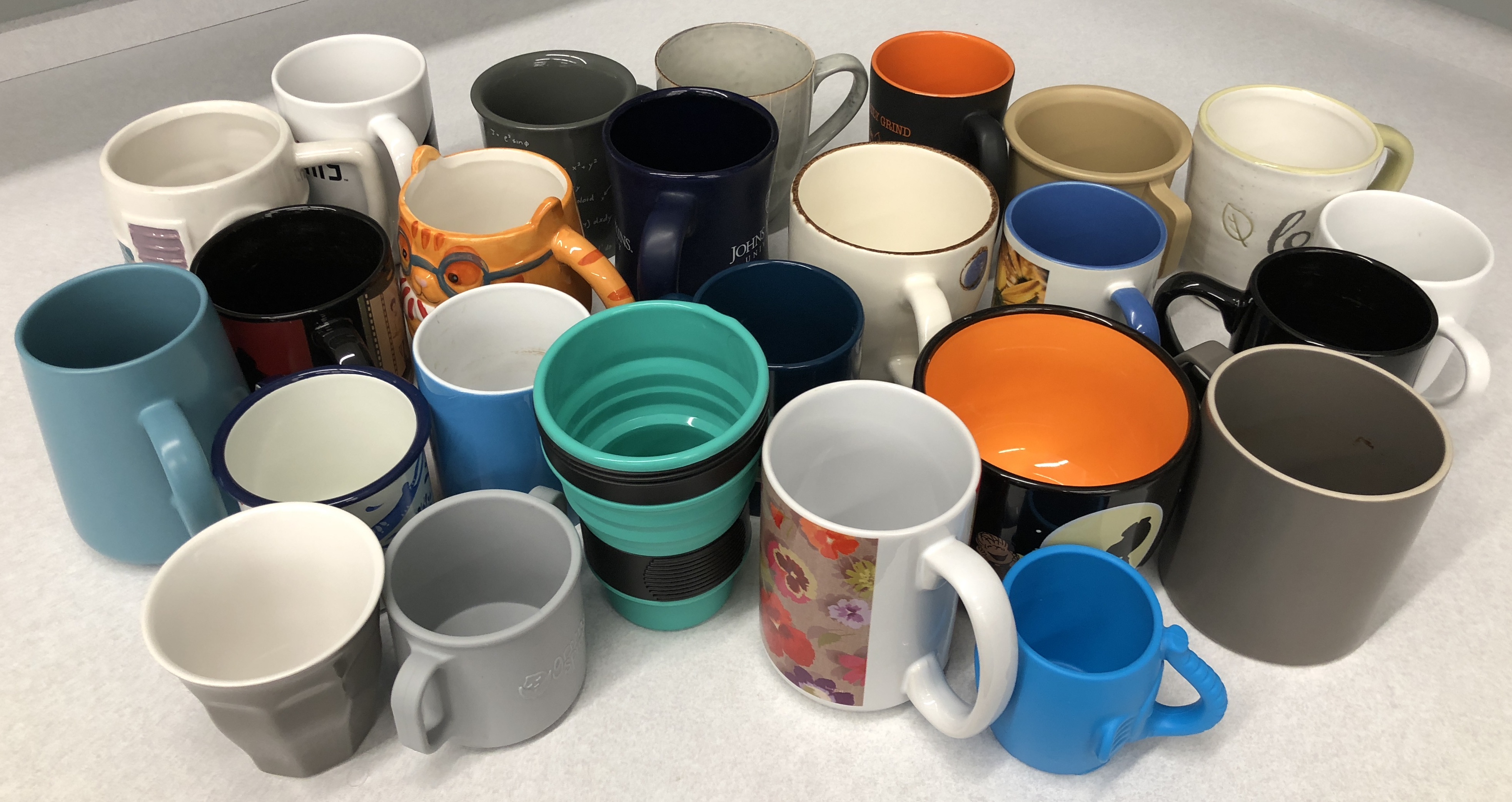}
\caption{All 25 mugs used in hardware experiments. Some of these mugs are very small or thin at the rim, raising challenges in perception using the real camera and requiring precise grasps.}
\label{fig:mugs}
\end{figure}

\begin{figure}[H]
\centering
\includegraphics[width=0.3\textwidth]{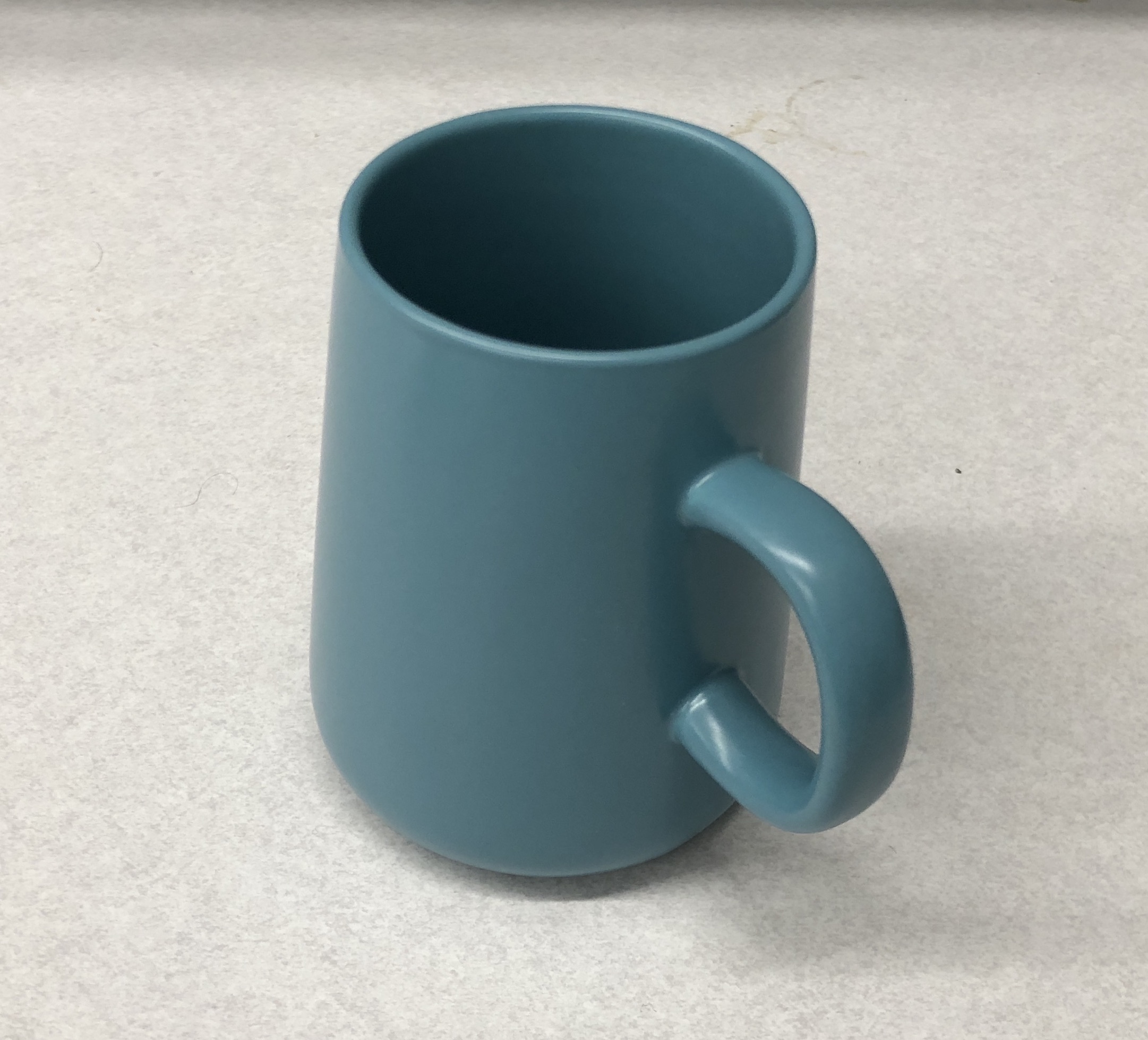}
\caption{The only mug failed to be grasped during one set of trials. It has a unique shape with a larger diameter at the bottom than at the rim.}
\label{fig:failed-mug}
\end{figure}

\begin{table}[H]
\scriptsize
\begin{center}
  \begin{tabular}{| c | c | c | c |}
    \hline
    No.\rule{0pt}{2ex} & Material & Dimensions (diameter and height, cm) \rule{0pt}{2ex} & Weight (g)  \rule{0pt}{2ex}\\
    \hline \hline
     1 \rule{0pt}{2ex} & Ceramic & 8.5, 11.6 \rule{0pt}{2ex} & 402.5 \rule{0pt}{2ex} \\  
     \hline
     2 \rule{0pt}{2ex} & Ceramic & 9.6, 11.9 \rule{0pt}{2ex} & 606.1 \rule{0pt}{2ex} \\  
     \hline
     3 \rule{0pt}{2ex} & Ceramic & 8.7, 8.8 \rule{0pt}{2ex} & 194.1 \rule{0pt}{2ex} \\ 
     \hline
     4 \rule{0pt}{2ex} & Ceramic & 8.3, 9.8 \rule{0pt}{2ex} & 302.1 \rule{0pt}{2ex} \\  
     \hline
     5 \rule{0pt}{2ex} & Ceramic &  8.3, 9.6 \rule{0pt}{2ex} & 337.5 \rule{0pt}{2ex} \\  
     \hline
     6 \rule{0pt}{2ex} & Ceramic & 9.7, 8.9 \rule{0pt}{2ex} & 423.6 \rule{0pt}{2ex} \\  
     \hline
     7 \rule{0pt}{2ex} & Ceramic & 9.6, 10.3 \rule{0pt}{2ex} & 530.3 \rule{0pt}{2ex} \\  
     \hline
     8 \rule{0pt}{2ex} & Ceramic & 8.5, 10.5 \rule{0pt}{2ex} & 349.0 \rule{0pt}{2ex} \\  
     \hline
     9 \rule{0pt}{2ex} & Ceramic & 9.6, 10.6 \rule{0pt}{2ex} & 453.8 \rule{0pt}{2ex} \\  
     \hline
     10 \rule{0pt}{2ex} & Ceramic & 8.7, 11.9 \rule{0pt}{2ex} & 426.6 \rule{0pt}{2ex} \\  
     \hline
     11 \rule{0pt}{2ex} & Ceramic & 9.7, 11.1 \rule{0pt}{2ex} & 360.3 \rule{0pt}{2ex} \\  
     \hline
     12 \rule{0pt}{2ex} & Ceramic & 11.2, 10.2 \rule{0pt}{2ex} & 443.6 \rule{0pt}{2ex} \\  
     \hline
     13 \rule{0pt}{2ex} & Ceramic & 8.3, 9.7 \rule{0pt}{2ex} & 350.3 \rule{0pt}{2ex} \\  
     \hline
     14 \rule{0pt}{2ex} & Ceramic & 8.2, 9,5 \rule{0pt}{2ex} & 351.1 \rule{0pt}{2ex} \\  
     \hline
     15 \rule{0pt}{2ex} & Ceramic & 12.7, 9.8 \rule{0pt}{2ex} & 426.6 \rule{0pt}{2ex} \\
     \hline
     16 \rule{0pt}{2ex} & Ceramic & 9.0, 10.7 \rule{0pt}{2ex} & 354.4 \rule{0pt}{2ex} \\  
     \hline
     17 \rule{0pt}{2ex} & Ceramic & 7.7, 11.9 \rule{0pt}{2ex} & 400.5 \rule{0pt}{2ex} \\  
     \hline
     18 \rule{0pt}{2ex} & Ceramic & 9.8, 11.5 \rule{0pt}{2ex} & 342.6 \rule{0pt}{2ex} \\  
     \hline
     19 \rule{0pt}{2ex} & Ceramic & 8.6, 11.7 \rule{0pt}{2ex} & 409.4 \rule{0pt}{2ex} \\  
     \hline
     20 \rule{0pt}{2ex} & Ceramic & 9.7, 10.9 \rule{0pt}{2ex} & 373.3 \rule{0pt}{2ex} \\  
     \hline
     21 \rule{0pt}{2ex} & Rubber & 6.8, 7.1 \rule{0pt}{2ex} & 85.7 \rule{0pt}{2ex} \\  
     \hline
     22 \rule{0pt}{2ex} & Rubber & 7.8, 7.2 \rule{0pt}{2ex} & 100.8 \rule{0pt}{2ex} \\  
     \hline
     23 \rule{0pt}{2ex} & Rubber & 9.3, 10.8 \rule{0pt}{2ex} & 124.0 \rule{0pt}{2ex} \\  
     \hline
     24 \rule{0pt}{2ex} & Stainless steel & 9.2, 8.1 \rule{0pt}{2ex} & 130.8 \rule{0pt}{2ex} \\  
     \hline
     25 \rule{0pt}{2ex} & Plastic & 10.6, 9.7 \rule{0pt}{2ex} & 171.4 \rule{0pt}{2ex} \\  
     \hline
  \end{tabular}
\vspace{5pt}
\caption{Materials, dimensions, and weights of all 25 mugs used in hardware experiments} 
\label{tab:mugs}
\end{center}
\end{table}
\vspace{-5pt}
\subsubsection{Pushing boxes}
\vspace{-5pt}
\begin{figure}[H]
\centering
\includegraphics[width=0.45\textwidth]{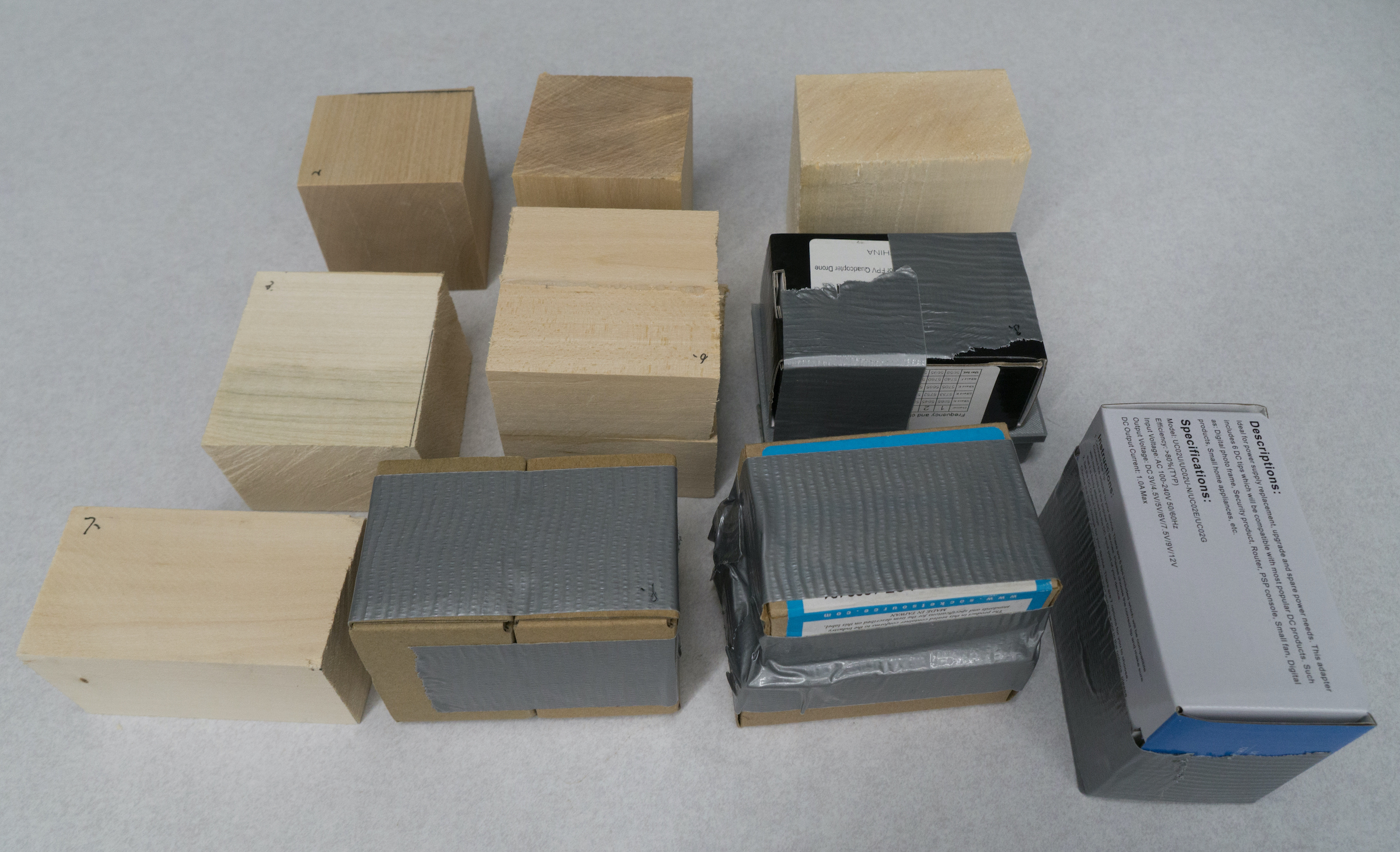}
\caption{All boxes used in hardware experiments. Some wood blocks are used in more than one trial as they can be placed in different orientations.}
\label{fig:boxes}
\end{figure}

\begin{table}[H]
\scriptsize
\begin{center}
  \begin{tabular}{| c | c | c | c |}
    \hline
    No.\rule{0pt}{2ex} & Material & Dimensions (cm) \rule{0pt}{2ex} & Weight (g)  \rule{0pt}{2ex}\\
    \hline \hline
     1 \rule{0pt}{2ex} & Wood & 7.2, 6.4, 6.4 \rule{0pt}{2ex} & 195.6 \rule{0pt}{2ex} \\  
     \hline
     2 \rule{0pt}{2ex} & Wood & 6.4, 7.2, 6.4 \rule{0pt}{2ex} & 195.6 \rule{0pt}{2ex} \\  
     \hline
     3 \rule{0pt}{2ex} & Wood & 6.4, 6.4, 7.2 \rule{0pt}{2ex} & 195.6 \rule{0pt}{2ex} \\  
     \hline
     4 \rule{0pt}{2ex} & Wood & 5.6, 6.4, 6.4 \rule{0pt}{2ex} & 145.2 \rule{0pt}{2ex} \\  
     \hline
     5 \rule{0pt}{2ex} & Wood & 6.4, 6.4, 5.6 \rule{0pt}{2ex} & 145.2 \rule{0pt}{2ex} \\  
     \hline
     6 \rule{0pt}{2ex} & Wood & 5.0, 8.0, 7.0 \rule{0pt}{2ex} & 120.5 \rule{0pt}{2ex} \\  
     \hline
     7 \rule{0pt}{2ex} & Wood & 5.0, 7.0, 8.0 \rule{0pt}{2ex} & 120.5 \rule{0pt}{2ex} \\  
     \hline
     8 \rule{0pt}{2ex} & Wood & 5.0, 9.0, 7.5 \rule{0pt}{2ex} & 143.3 \rule{0pt}{2ex} \\  
     \hline
     9 \rule{0pt}{2ex} & Wood & 5.0, 7.5, 9.0 \rule{0pt}{2ex} & 143.3 \rule{0pt}{2ex} \\  
     \hline
     10 \rule{0pt}{2ex} & Wood &7.5, 8.0, 7.5 \rule{0pt}{2ex} & 167.8 \rule{0pt}{2ex} \\  
     \hline
     11 \rule{0pt}{2ex} & Wood & 5.0, 8.0, 5.0 \rule{0pt}{2ex} & 126.7 \rule{0pt}{2ex} \\  
     \hline
     12 \rule{0pt}{2ex} & Cardboard & 10.0, 6.0, 6.0 \rule{0pt}{2ex} & 105.5 \rule{0pt}{2ex} \\  
     \hline
     13 \rule{0pt}{2ex} & Cardboard & 5.8, 9.0, 7.3 \rule{0pt}{2ex} & 331.8 \rule{0pt}{2ex} \\  
     \hline
     14 \rule{0pt}{2ex} & Cardboard & 6.5, 9.2, 7.5 \rule{0pt}{2ex} & 382.1 \rule{0pt}{2ex} \\  
     \hline
     15 \rule{0pt}{2ex} & Cardboard & 5.4, 10.4, 7.4 \rule{0pt}{2ex} & 226.1 \rule{0pt}{2ex} \\  
     \hline
  \end{tabular}
\vspace{5pt}
\caption{Materials, dimensions, and weights of all 15 rectangular boxes used in hardware experiments. Cardboard boxes are filled with weights. Some of the entries have the same weight since they refer to the same box with different orientations.} 
\label{tab:pushing-boxes}
\end{center}
\end{table}


\end{document}